\documentclass[11pt,twoside]{article} 
\usepackage{amsmath,amsfonts,bm}

\def\eqref#1{equation~\ref{#1}}

\def\1{\bm{1}}

\def\mH{{\bm{H}}}

\def\mX{{\bm{X}}}
\def\mY{{\bm{Y}}}

\DeclareMathAlphabet{\mathsfit}{\encodingdefault}{\sfdefault}{m}{sl}
\SetMathAlphabet{\mathsfit}{bold}{\encodingdefault}{\sfdefault}{bx}{n}

\usepackage{amsmath,amsfonts,bm,amsthm}

\newtheorem{assumption}{Assumption}

\newtheorem{definition}{Definition}

\newcommand{\natset}[1]{\llbracket {#1} \rrbracket}
\newcommand{\intset}[1]{\{ 0, 1, \ldots, {#1} \}}
\newcommand{\ones}[1]{\mathbbm{1}_{#1}}
\newcommand{\identity}[1]{\mathbb{I}_{#1}}

\newcommand{\norm}[1]{\left\lVert#1\right\rVert}

\def\RR{\mathbb{R}}

\def\Ff{\mathcal{F}}

\usepackage[utf8]{inputenc} 
\usepackage[T1]{fontenc}    
\usepackage{mathtools}
\usepackage{stmaryrd}
\usepackage{enumitem}
\usepackage{booktabs}       
\usepackage{amsfonts}       
\usepackage{nicefrac}       
\usepackage{microtype}      
\usepackage{subfigure}
\usepackage{epsf}
\usepackage{epsfig}
\usepackage{fancyhdr}
\usepackage{graphics}
\usepackage{graphicx}
\usepackage{psfrag}
\usepackage{fullpage}
\usepackage{pdfpages}
\usepackage{url}
\usepackage[colorlinks,linkcolor=magenta,citecolor=blue,pagebackref=true]{hyperref}
\renewcommand*{\backrefalt}[4]{%
    \ifcase #1 \footnotesize{(Not cited.)}%
    \or        \footnotesize{(Cited on page~#2.)}%
    \else      \footnotesize{(Cited on pages~#2.)}%
    \fi}

\newcommand{\widgraph}[2]{\includegraphics[keepaspectratio,width=#1]{#2}}

\newcommand{\step}{\text{step}}
\usepackage{color}
\usepackage{amsthm}
\usepackage{amsmath}
\usepackage{amssymb,bbm}
\usepackage{caption}
\usepackage{algorithmic}
\usepackage{algorithm}
\usepackage{multirow}
\usepackage{multicol}
\usepackage{tablefootnote}
\usepackage{tabu}

\begin{document}
\begin{center}
    {\bf{\LARGE{On Cross-Layer Alignment for Model Fusion of Heterogeneous Neural Networks}}}
      
    \vspace*{.2in}
    
    {\large{
        \begin{tabular}{cccccc}
            Dang Nguyen$^{\diamond}$ & Trang Nguyen$^{\star}$ & Khai Nguyen$^{\dagger}$ & Dinh Phung$^{\ddagger}$ & Hung Bui$^{\diamond}$ & Nhat Ho$^{\dagger}$
        \end{tabular}
    }}
    
    \vspace*{.2in}
    
    \begin{tabular}{ccc}
        VinAI Research$^{\diamond}$ & Hanoi University of Science and Technology$^{\star}$ & University of Texas, Austin$^{\dagger}$ \\ 
        & Monash University$^{\ddagger}$ & \\
    \end{tabular}
    
    
    \vspace*{.2in}
    
    \begin{abstract}
        Layer-wise model fusion via optimal transport, named OTFusion, applies soft neuron association for unifying different pre-trained networks to save computational resources. While enjoying its success, OTFusion requires the input networks to have the same number of layers. To address this issue, we propose a novel model fusion framework, named \emph{CLAFusion}, to fuse neural networks with a different number of layers, which we refer to as heterogeneous neural networks, via cross-layer alignment. The cross-layer alignment problem, which is an unbalanced assignment problem, can be solved efficiently using dynamic programming. Based on the cross-layer alignment, our framework balances the number of layers of neural networks before applying layer-wise model fusion. Our experiments indicate that CLAFusion, with an extra finetuning process, improves the accuracy of residual networks on the CIFAR10, CIFAR100, and Tiny-ImageNet datasets. Furthermore, we explore its practical usage for model compression and knowledge distillation when applying to the teacher-student setting.
    \end{abstract}
\end{center}

\section{Introduction}
\label{sec:introduction}
The ubiquitous presence of deep neural networks raises the question, ``Can we combine the knowledge of two and more deep neural networks to improve performance?" As one of the earliest successful methods, ensemble learning aggregates the output over a collection of trained models, which is referred to as an ensemble, to improve the generalization ability~\cite{hansen1990neural,zhou2012ensemble}. However, it is expensive in terms of computational resources because it requires storing and running many trained neural networks during inference. To overcome the previous hardness, we need to find a single neural network that can inherit the knowledge from multiple pre-trained neural networks and have a small size at the same time. To our best knowledge, there are two popular approaches for this purpose: knowledge distillation and model fusion.  

\vspace{0.5em}\noindent
The first approach, \emph{knowledge distillation}, is a machine learning technique that transfers knowledge from large networks (teachers) into a smaller one (student)~\cite{hinton2015distilling}. The smaller network is trained by the task-specific loss and additional distillation losses that encourage the student network to mimic the prediction of the teachers. Knowledge distillation is well-known as an efficient model compression and acceleration technique~\cite{gou2021distilling}. The distillation process, however, is computationally expensive because it runs forward inference for both teacher and student networks at each training epoch. In addition, simply distilling into a randomly initialized student network, in most cases, does not yield high performance, thus a good initialization of the student network is needed~\cite{turc2019well}.

\vspace{0.5em}\noindent
The second line of work is called \emph{model fusion} which is the problem of merging a collection of pre-trained networks into a unified network~\cite{wang2020federated,singh2020model}. The simplest technique in model fusion is vanilla averaging~\cite{utans1996weight,smith2017investigation}, which computes the weighted average of pre-trained network parameters without the need for retraining. Their methods fuse neural networks with the same architecture without considering the permutation invariance nature of neural networks. To mitigate this problem, the idea of solving a neuron alignment problem before applying weight averaging is utilized~\cite{ashmore2015method,li2015convergent}. Recently, there are two concurrent works that benefit from the idea of neuron alignment. FedMA~\cite{wang2020federated} formulated and solved the assignment problem to find a permutation matrix that undoes the permutation of weight matrices. While OTFusion~\cite{singh2020model} viewed the problem through the lens of optimal transport~\cite{monge1781memoire,kantorovich2006translocation} and used the transport map to align the weight matrices. Though both methods can fuse multiple neural networks, their applications are limited to networks with the same number of layers.

\vspace{0.5em}\noindent
\textbf{Contributions.} In this paper, we propose a model fusion framework for fusing heterogeneous (\emph{unequal width and unequal depth}) neural networks, which is named as \textit{Cross-Layer Alignment Fusion} (CLAFusion). CLAFusion consists of three parts: cross-layer alignment, layer balancing method, and layer-wise model fusion method. In summary, our main contributions are two-fold:

\vspace{0.5em}\noindent
1. We formulate the cross-layer alignment as an assignment problem using layer representation and layer similarity, then propose a dynamic programming-based algorithm to solve it. Next, we present two natural and fast methods to balance the number of layers between two networks. Furthermore, we discuss the application of our framework for the cases of more than two networks.

\vspace{0.5em}\noindent
2. Our experiments demonstrate the efficiency of CLAFusion on different setups. The fused model from CLAFusion serves as an efficient initialization when training residual networks. In addition, the components of CLAFusion can be successfully applied to the heterogeneous model transfer task. Moreover, CLAFusion shows potential applications for model compression and knowledge distillation in the teacher-student setting. 

\vspace{0.5em}\noindent
\textbf{Organization.} The rest of the paper is organized as follows. After reviewing some background in Section~\ref{sec:background}, we introduce CLAFusion in Section~\ref{sec:framework}. We study their performance in Section~\ref{sec:experiments} on the image classification task with various settings and followed by discussions in Section~\ref{sec:conclusion}. Finally, experimental settings and additional experiments are deferred to the supplementary material.

\section{Background}
\label{sec:background}
In this section, we first recall the layer representation and layer similarity that have been widely used in applications. Then, we review available model fusion methods and their limitations. Finally, we discuss OTFusion and the challenge of fusing heterogeneous neural networks.

\subsection{Layer Representation and Layer Similarity}
\label{subsec:layer_representation}
A common layer representation is the matrix of activations~\cite{kornblith2019similarity}. Activation matrix, also known as activation map, can be used as a type of knowledge to guide the training in knowledge distillation~\cite{gou2021distilling}. Another choice is the weight matrix of the trained neural networks, which has been shown to perform well on intra-network comparison tasks~\cite{o2021layer}.

\vspace{0.5em}\noindent
The neural network exhibits the same output when its neurons in each layer are permuted, this is so-called the permutation invariance nature of neural network parameters. Due to the permutation invariance property, a meaningful layer similarity index should be invariant to orthogonal transformations~\cite{kornblith2019similarity}. Their proposed Centered Kernel Alignment (CKA) effectively identifies the relationship between layers of different architectures.

\subsection{Model fusion}
\label{subsec:model_fusion}
Despite showing good empirical results, vanilla averaging~\cite{utans1996weight,smith2017investigation} only works in the case when the weights of individual networks are relatively close in the weight space. As an effective remedy, several works on model fusion considered performing permutation of neurons in hidden layers before applying vanilla averaging. FBA-Wagging~\cite{ashmore2015method}, Elastic Weight Consolidation~\cite{Leontev_2019}, and FedMA~\cite{wang2020federated} formulate the neuron association problem as an assignment problem and align the neurons. Nevertheless, those variants are not generalized to heterogeneous neural networks. 

\vspace{0.5em}\noindent
There is limited literature devoted to the discussion of fusing heterogeneous neural networks. NeuralMerger~\cite{chou2018unifying} is one of the few attempts to deal with heterogeneous neural networks. However, there are two key differences in NeuralMerger from our CLAFusion. First, their cross-layer alignment is hand-crafted and dedicated to specific architectures. Secondly, they decompose weights into lower dimensions and use vector quantization to merge the weights. On the other hand, we introduce a systematic way to solve the cross-layer alignment problem, and combining weights is done using the layer-wise model fusion method.

\subsection{Model fusion via Optimal Transport}
\label{subsec:otfusion}
Recently, Singh et al.~\cite{singh2020model} proposed OTFusion, which solves the neuron association problem based on free-support barycenters~\cite{cuturi2014fast}. OTFusion leads to a noticeable improvement over vanilla averaging when all individual networks have the same architecture. One advantage of OTFusion over other vanilla averaging-based model fusion methods is that it can fuse networks with different widths (i.e., number of neurons) thanks to the nature of optimal transport.

\vspace{0.5em}\noindent
\textbf{Issues of OTFusion.} OTFusion, however, is still a layer-wise approach that cannot be directly employed for heterogeneous neural networks. We need to find two corresponding layers, matching their neurons before averaging weight matrices. There are two challenges to this scheme. Firstly, the one-to-one mapping between layers is not available in advance. A naive one-to-one mapping of layers with the same index as the  layer-wise approach may cause a mismatch when the numbers of layers are different. For example, we analyze two VGG configurations~\cite[Table~1]{simonyan2014very} with different depths: VGG11 and VGG13. The second convolution layer of VGG11 has $128$ channels and an input image of size $112 \times 112$. While that of VGG13 has $64$ channels and an input image of size $224 \times 224$. Secondly, assume that we have successfully found a one-to-one mapping, there still necessitates a special treatment for remaining layers that has no counterparts. Simply removing those layers reduces the performance of the deeper network, this may cumulatively degrade the fused model.

\begin{figure*}[t!]
    \begin{center}
        \begin{tabular}{c}
            \widgraph{0.95\textwidth}{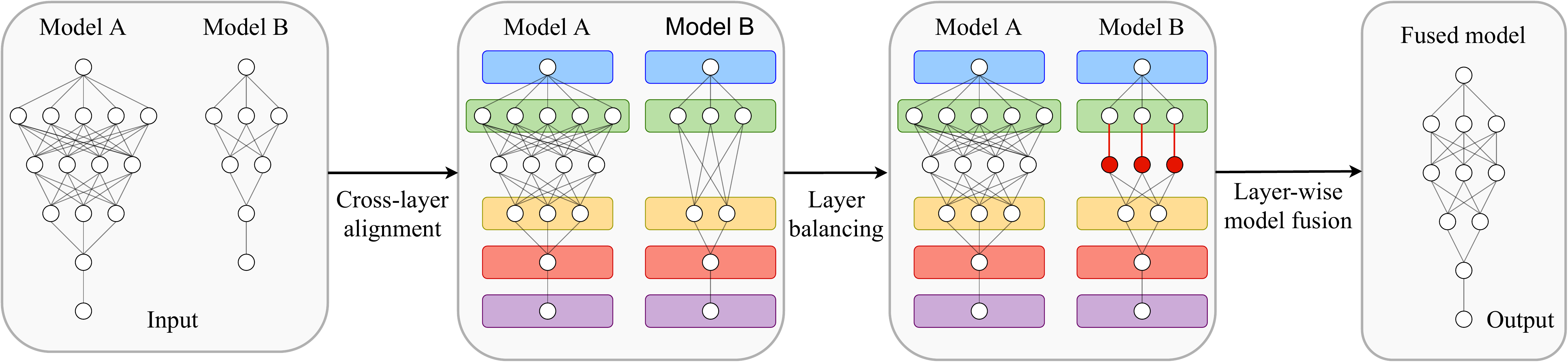}
        \end{tabular}
    \end{center}
    \caption{
        \footnotesize{\textbf{CLAFusion strategy: } In the first step, the cross-layer alignment problem is solved for two pre-trained models. Two corresponding layers are surrounded by two rounded rectangles of the same color. Based on the optimal mapping obtained in the previous step, CLAFusion balances the number of layers (adding layers in this figure). The red circles and lines indicate the newly added neurons and weights (Zero weights are omitted). Finally, CLAFusion applies a layer-wise model fusion method to produce the final output.}} 
    \label{fig:framework}
\end{figure*}

\section{Cross-layer Alignment Model Fusion}
\label{sec:framework}
To overcome the challenges of OTFusion, in this section, we propose Cross-layer Alignment Model Fusion (in short, \emph{CLAFusion}) framework for fusing heterogeneous neural networks. 
In this section, we first provide an overview of our framework, followed by the formal definition and an efficient solution to the CLA problem. Next, we introduce two layer balancing methods that will be in our experiments. Finally, we describe a simple yet effective extension of our framework to the case of multiple neural networks.

\vspace{0.5em}\noindent
\textbf{Notation.} We define $\natset{n}$ as a set of integers $\{ 1, \ldots, n \}$ and $\identity{n}$ as an identity matrix of size $n \times n$. We use A, B to denote the individual models and $\Ff$ to denote the fused model. $l$ indicates the layer index. Layer $l$ of model A has $p_A^{(l)}$ neurons and an activation function $f_A^{(l)}$. The pre-activation and activation vector at layer $l$ of model A are denoted as $\boldsymbol{z}_A^{(l)}, \boldsymbol{x}_A^{(l)} \in \RR^{p_A^{(l)}}$, respectively. The weight matrix between layers $l$ and $l-1$ of model A is $\boldsymbol{W}_A^{(l, l-1)}$.  The following equations hold between two consecutive layers of model A (we omit the bias terms here).

\vspace{-0.5em}
\begin{align} \label{eq:feed_forward}
    \boldsymbol{x}^{(l)} = f_A^{(l)}(\boldsymbol{z}_A^{(l)}) =  f_A^{(l)}(\boldsymbol{W}_A^{(l, l-1)} \boldsymbol{x}^{(l-1)}).
\end{align}

\noindent
We stack the pre-activation vector over $t$ samples to form a $t-$row pre-activation matrix. Let $\boldsymbol{Z}_A^{(l)} \in \RR^{t \times p_A^{(l)}}$ denote a matrix of pre-activations at layer $l$ of model A for $t$ samples, and $\boldsymbol{Z}_B^{(l)} \in \RR^{t \times p_B^{(l)}}$ denote a matrix of pre-activations at layer $l$ of model B for the same $t$ samples. The activation matrices $\boldsymbol{X}_A^{(l)}, \boldsymbol{X}_B^{(l)}$ are obtained by applying the corresponding activation functions element-wise to the pre-activation matrices.

\vspace{0.5em}\noindent
\textbf{Problem setting.} Hereafter, we consider fusing two feed-forward networks A and B of the \emph{same architecture family}. Two networks have the same input and output dimension but a different number of hidden layers. In each network, the hidden layer has an activation function of ReLU, which is the general setting in modern architectures. Additional work is required to handle bias terms and the batch normalization layer properly so we leave them for future work. Let $m$ and $n$ be the number of hidden layers of models A and B, respectively. Taking the input and output layers into account, the numbers of layers are $m + 2$ and $n + 2$, respectively. Without loss of generality, assume that $m \geq n$. The layer index of model A and B are $\intset{m+1}$ and $\intset{n+1}$, respectively.

\vspace{0.5em}\noindent
\textbf{General strategy.} Our framework has three components, which are illustrated in Figure~\ref{fig:framework}. The first part is a cross-layer alignment which is a one-to-one mapping from hidden layers of model B to hidden layers of model A. The second part is a layer balancing method that equalizes the number of layers of two models based on the cross-layer alignment. The last part is a layer-wise model fusion method that is applied to same-sized models. The third part of our framework adopts any model fusion methods that can fuse two neural networks with the same number of layers.

\begin{figure*}[t!]
    \begin{center}
        \begin{tabular}{c}
            \widgraph{0.8\textwidth}{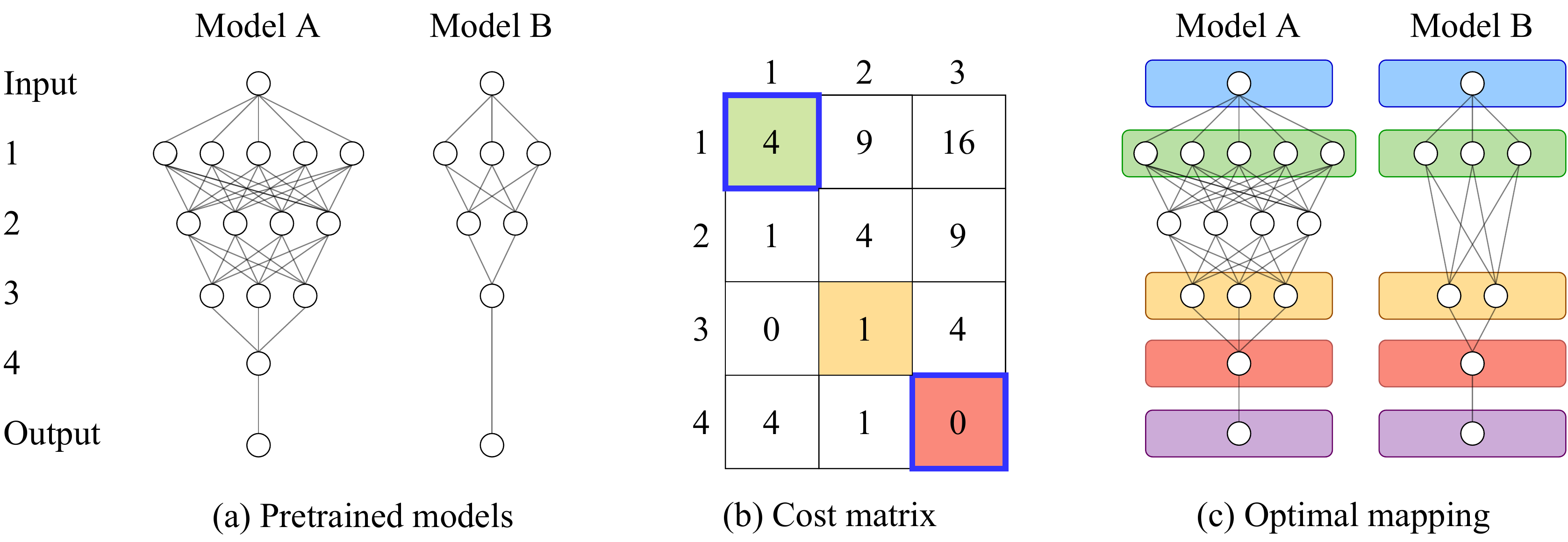} 
        \end{tabular}
    \end{center}
    \caption{
        \footnotesize{\textbf{Cross-layer alignment example: } Two pre-trained neural networks are given in (a). Model A has 4 hidden layers of size 5, 4, 3, 1 while model B has 3 hidden layers of size 3, 2, 1. (b) shows the cost matrix (squared difference) between layer representations (number of neurons) for hidden layers of two networks. Three color cells represent the solution of the CLA problem. Note that the upper-left and lower-right cells are automatically chosen by the constraints on the first and last hidden layers. (c) visualizes the optimal mapping. Two rounded rectangles of the same colors represent two corresponding layers in the optimal CLA.}} 
    \label{fig:cross_layer_alignment}
\end{figure*}

\subsection{Cross-layer alignment}
\label{subsec:cross_layer_alignment}
Cross-layer alignment (CLA) is a nontrivial problem that also arises in the context of the feature-based knowledge distillation approach~\cite{zagoruyko2016paying,tung2019similarity,chen2021cross}. CLA requires finding a one-to-one function that matches similar layers between two models. Despite considering CLA, prior works are based on the hand-crafted layer association, which causes a lack of generalization. To address this matter, we introduce an efficient method for solving the CLA problem. Because the input and output layers of the two models are identical in both dimension and functionality, we only align their hidden layers. In addition, the first and last hidden layers usually play a critical role in the model performance~\cite{zhang2019all}. Therefore, we directly match the first and last hidden layers of the two models. In case $n = 1$, we prioritize the first layer over the last layer. The importance of this constraint in CLA is studied in Appendix~\ref{sec:ablation_studies}. Furthermore, two networks are feed-forward, so the mapping is necessary to keep the sequential order of layers. All in all, we formulate the CLA problem as follows. 

\begin{definition}
\label{def:cross_layer_alignment}
    Assume that we have two layer representations for hidden layers of two models as, $\boldsymbol{L}_A = \{ L_A^{(1)}, \ldots, L_A^{(m)}\}$ and $\boldsymbol{L}_B = \{ L_B^{(1)}, \ldots, L_B^{(n)}\}$. Let $\boldsymbol{C} \in \RR^{m \times n}$ denote the cost matrix between them, i.e., $\boldsymbol{C}_{i, j} = d(L_A^{(i)}, L_B^{(j)}), i \in \natset{m}, j \in \natset{n}$ where $d$ is a layer dissimilarity. The \emph{optimal CLA} is a strictly increasing mapping $a : \natset{n} \mapsto \natset{m}$ satisfying $a(1) = 1, a(n) = m$ and minimizing $S(n, m) = \sum_{i=1}^{n} \boldsymbol{C}_{a(i), i}$.
\end{definition}

\noindent
\textbf{Layer representation.} Empirically, we used the pre-activation matrix, which is also used in the activation-based alignment strategy of OTFusion~\cite[Section~4]{singh2020model}, as the layer representation. To measure the discrepancy between two pre-activation matrices, we choose the dissimilarity index based on Linear CKA~\cite{kornblith2019similarity}. Specifically, given two pre-activation matrices $\mX \in \RR^{t \times p_A}$ and $\mY \in \RR^{t \times p_B}$ for the same $t$ samples of two models A and B, their centered versions are defined as $\bm{\hat{X}} = \mX \mH_{p_A}$ and $\bm{\hat{Y}} = \mY \mH_{p_B}$ where $\mH_{p} = \identity{p} - \frac{1}{n} \ones{p} \ones{p}^T$. Then, the similarity index Linear CKA between $\mX$ and $\mY$ has the following form
\begin{equation}
    \frac{\norm{\bm{\hat{Y}}^T \bm{\hat{X}}}_F^2}{\norm{\bm{\hat{X}}^T \bm{\hat{X}}}_F \norm{\bm{\hat{Y}}^T \bm{\hat{Y}}}_F}
\end{equation}
where $\norm{\cdot}_F$ is the Frobenius norm. Then, the layer dissimilarity $d$ in Definition~\ref{def:cross_layer_alignment} is empirically set to 
\begin{equation}
    1 - \frac{\norm{\bm{\hat{Y}}^T \bm{\hat{X}}}_F^2}{\norm{\bm{\hat{X}}^T \bm{\hat{X}}}_F \norm{\bm{\hat{Y}}^T \bm{\hat{Y}}}_F}.
\end{equation}

\noindent
It is worth noting that besides the above option, our framework can work with any appropriate alternatives, such as the number of neurons for layer representation and Euclidean distance for dissimilarity.

\vspace{0.5em}\noindent
\textbf{Toy example.} Figure~\ref{fig:cross_layer_alignment} presents a CLA problem, in which two models in Figure~\ref{fig:framework} are utilized again. Here, the layer representation is the number of neurons with the squared difference as the cost function. The optimal mapping $a$ obtained from solving the CLA problem is $a(1)=1, a(2)=3, a(3)=4$.

\vspace{0.5em}\noindent
\textbf{Discussion.} The CLA problem in Definition~\ref{def:cross_layer_alignment} is a special case of the (linear) unbalanced assignment problem~\cite{ramshaw2012minimum}. The condition of strictly increasing ensures the sequential order of layers when matching. If we have two increasing layer representations in 1-D, the problem can be interpreted as the Partial Optimal Transport in 1-D problem with uniform weights. It can be solved efficiently in $\mathcal{O}(mn)$ using the proposed method in Bonneel et al.~\cite{bonneel19SPOT}. However, the increasing constraint on layer representations is quite strict and layer representations might not be in 1-D in general. On the other hand, the unbalanced assignment problem can be solved using the generalization of the Hungarian algorithm~\cite{kuhn1955hungarian}. The time complexity of the Hungarian algorithm in our CLA problem is $\mathcal{O}(mn^2)$. The importance of CLA in our framework is demonstrated in Appendix~\ref{sec:ablation_studies} where we compare the performance of the fused model obtained using the optimal mapping with those from other mappings.

\vspace{0.5em}\noindent
\textbf{Cross-layer alignment algorithm.} We propose an efficient algorithm based on dynamic programming to solve the CLA problem. Algorithm~\ref{alg:layer_alignment} details the pseudo-code for our algorithm. Given the cost matrix between layer representations of two networks, we want to find the optimal alignment from the shallower network (model B) to the deeper network (model A). The optimal alignment is a strictly increasing mapping that minimizes the total assignment cost. We denote $S(i, j)$ as the optimal assignment cost from the first $i$ layers of model B to the first $j$ layers of model A. We initialize $n \times m$ entities in matrix $S$ to be $0$. Due to the strictly increasing property, when $i = j$, the optimal alignment is trivial. Assume that we have already found all optimal assignment costs before $S(i, j) (i <= j)$. There are two situations. If layer $j$ of model A is mapped to one layer in model B. Because the optimal alignment is an increasing mapping, layer $j$ of model A must be assigned to layer $i$ of model B. Otherwise, we can find another mapping that has a smaller assignment cost. The optimal assignment cost in this situation is $S(i-1, j-1) + C_{j, i}$. In the second situation, layer $j$ of model A is not mapped to any layers of model B. Thus, the optimal assignment cost is $S(i, j-1)$. Considering all scenarios, the optimal assignment cost is the lower cost between $S(i-1, j-1) + C_{j, i}$ and $S(i, j-1)$. After finding the optimal assignment cost $S(n, m)$ we can backtrack in the reverse direction to get the optimal CLA mapping by finding pairs $(i, j)$ such that $S(i, j) = S(i-1, j-1) + C_{j, i}$. Note that we have two constraints on the first and last hidden layers. Therefore, $a(1) \leftarrow 1; a(n) \leftarrow m$.

\setlength{\textfloatsep}{0.1in}
\begin{algorithm}[t!]
    \caption{Cross-layer alignment algorithm}
    \label{alg:layer_alignment}
    \begin{algorithmic}
        \STATE {\bfseries Input:} $\boldsymbol{C} = \left[ \boldsymbol{C}_{i, j} \right]_{i, j}$
        \STATE $S(i, i) \leftarrow \sum_{l=1}^{i} \boldsymbol{C}_{l, l}, i \in \natset{n}$
        \STATE $S(0, j) \leftarrow 0, j \in \natset{m}$ and $S(i, 0) \leftarrow 0, i \in \natset{n}$
        \FOR{$i=1$ {\bfseries to} $n$}
        \FOR{$j=i+1$ {\bfseries to} $m$}
        \STATE $S(i, j) \leftarrow \min \{ S(i, j-1), S(i-1, j-1) + \boldsymbol{C}_{j, i} \}$
        \ENDFOR
        \ENDFOR
        \STATE $a(1) \leftarrow 1; a(n) \leftarrow m; i \leftarrow n-1; j \leftarrow m-1$
        \WHILE{$i \geq 2$}
        \WHILE{$j \geq i+1$ and $S(i, j) = S(i, j-1)$}
        \STATE $j \leftarrow j - 1$
        \ENDWHILE
        \STATE $a(i) = j; i \leftarrow i - 1; j \leftarrow j - 1$
        \ENDWHILE
        \STATE {\bfseries Output:} $a$
    \end{algorithmic}
    \end{algorithm}
\setlength{\floatsep}{0.1in}

\vspace{0.5em}\noindent
Given the cost matrix $\boldsymbol{C}$, the time complexity of our algorithm is only $\mathcal{O}(mn)$. Note that the optimal alignment from Definition~\ref{def:cross_layer_alignment} is not necessarily unique so we choose one which is obtained by backtracking.

\vspace{0.5em}\noindent
\textbf{CLA for Convolutional neural network (CNN).} Following NeuralMerger, we do not match fully connected layers and convolution layers. Naturally, a convolution layer has different functionality from a fully connected layer. In addition, there is a lack of an appropriate way to combine the weights of a fully connected layer and a convolution layer. In this paper, we consider CNN architecture which consists of consecutive convolution layers followed by fully connected layers such as VGG~\cite{simonyan2014very} and RESNET~\cite{he2016deep}. The key idea is to solve the CLA for same-type layers separately then combine two mappings. We further break the convolution part into smaller groups and find the mapping for each pair of groups. In particular, we divide into 5 groups separated by the max-pooling layers in VGG. In RESNET, we group consecutive blocks with the same number of channels, a.k.a. stage, and in each stage, align blocks instead of layers. For the block representation, we choose the (pre-)activation matrix of the second convolution layer in each block.

\begin{figure*}[t]
    \begin{center}
        \begin{tabular}{c}
            \widgraph{0.6\textwidth}{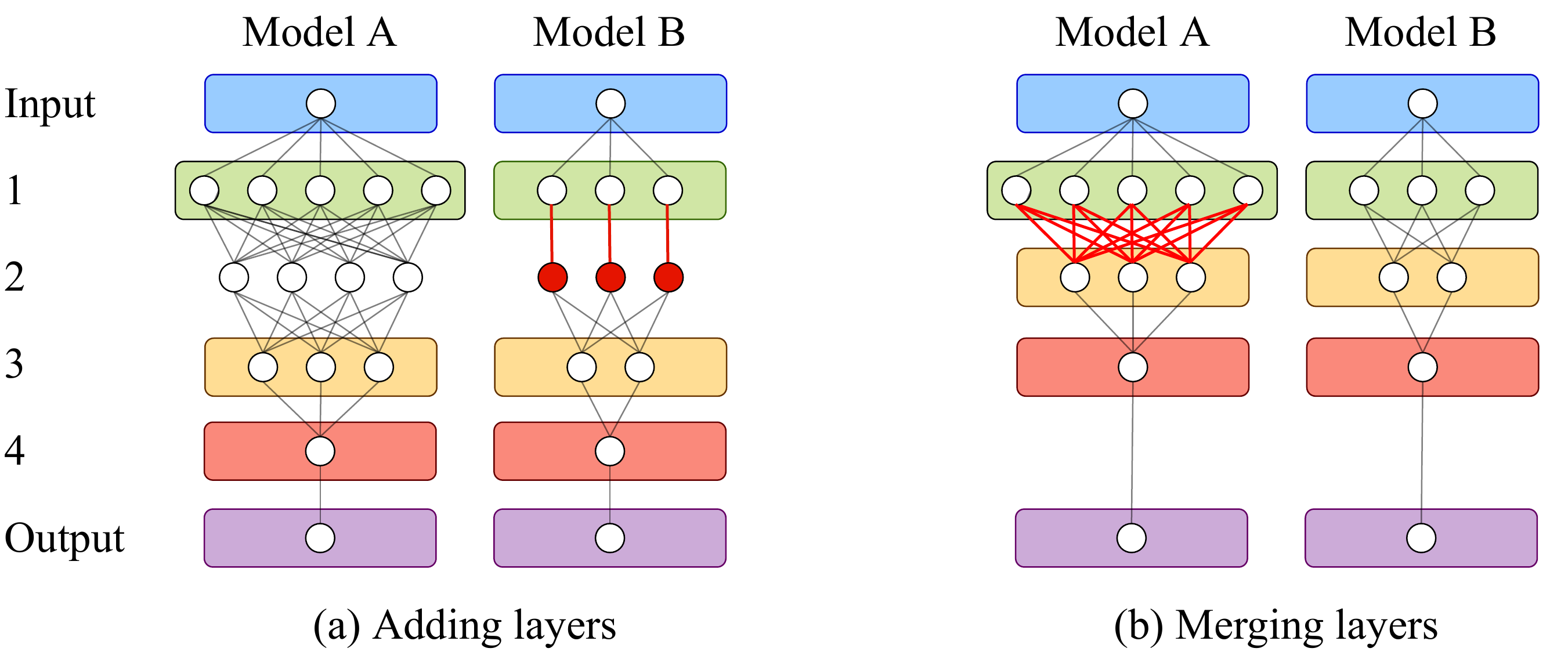} 
        \end{tabular}
    \end{center}
    \caption{
        \footnotesize{\textbf{Layer balancing examples: } In (a), a new layer is added between hidden layers 1 and 2 of model B. Newly added neurons and weights are in red (We omit zero weights). The new weight matrix between layers 1 and 2 of model B is an identity matrix whose size equals the number of neurons at layer 1. While the old weight matrix between layers 1 and 2 in model B becomes the new weight matrix between layers 2 and 3. In (b), hidden layer 2 of model A is removed. The red lines indicate the new weight matrix which is calculated based on two old weight matrices and the activation matrix of the removed layer.}} 
    \label{fig:layer_balancing}
\end{figure*} 

\subsection{Layer balancing methods}
\label{subsec:layer_balancing}
As mentioned above, some layers in model A may have no counterparts in model B in the optimal mapping. Naturally, we have two opposite directions: reduce layers in model A or add layers into model B. Assume that we have already balanced the number of layers up to layer $l$ of model B. At layer $l+1$ of model B, there are two possibilities. If $a(l+1) - a(l) = 1$, the layer-wise fusion method is ready to apply up to layer $l+1$. If $a(l+1) - a(l) > 1$, we can either merge layers between layers $a(l)$ and $a(l+1)$ of model A or add layers between layers $l$ and $l+1$ of model B. Next, we discuss two methods in the case $a(l+1) - a(l) = 2$. When $a(l+1) - a(l) > 2$, we can repeat the same method $a(l+1) - a(l) - 1$ times.

\vspace{0.5em}\noindent
\textbf{Add layers into model B.} We add a layer $l'$ between layers $l$ and $l+1$ of model B. The new weight matrices are defined as $\boldsymbol{W}_B^{(l', l)} \leftarrow \identity{p_B^{(l)}}$ and $\boldsymbol{W}_B^{(l+1, l')} \leftarrow \boldsymbol{W}_B^{(l+1, l)} \in \RR^{p_B^{(l+1)} \times p_B^{(l)}}$. The new activation function is defined as $f_B^{(l')} \leftarrow f_A^{(a(l) + 1)}$, which is ReLU. Because $x_B^{(l)} \succeq 0 \text{ for all } l \in \natset{n}$, from Equation~\ref{eq:feed_forward} we have

\begin{equation*}
    x_B^{(l')} = f_B^{(l')}(\boldsymbol{W}_B^{(l', l)} x_B^{(l)}) = ReLU(x_B^{(l)}) = x_B^{(l)}.
\end{equation*}

\noindent
Therefore, the information of model B remains unchanged after adding layer $l'$. Note that the new layer is just an identity mapping, which is a trick that has been used in RESNET and NET2NET~\cite{chen2015net2net}. Also discussed in NET2NET, the adding layers method is a function-preserving transformation that allows us to generate a valuable initialization for training a larger network. We later examine this hypothesis in our experiments.

\vspace{0.5em}\noindent
\textbf{Merge layers in model A.} We merge layer $a(l) + 1$ into layer $a(l)$ of model A by directly connecting layer $a(l)$ to layer $a(l+1)$. Because $f_A^{(a(l)+1)}$ is ReLU, the new weight matrix $\boldsymbol{W}_A^{(a(l+1), a(l))}$ can be written as
\begin{equation*}
    \boldsymbol{W}_A^{(a(l+1), a(l))} = \boldsymbol{W}_A^{(a(l+1), a(l)+1)} \boldsymbol{D}_A^{(a(l)+1)} \boldsymbol{W}_A^{(a(l)+1, a(l))},
\end{equation*}
where $\boldsymbol{D}_A^{(a(l)+1)} \in \RR^{p_A^{(a(l)+1)} \times p_A^{(a(l)+1)}}$ is an \emph{input-dependent} diagonal matrix with 0s and 1s on its diagonal. We denote the sign (binary step) function as follow $\step(x) = 1$ if $x >= 0$ and $\step(x) = 0$ if $x < 0$. Then, the $i^{th}$ entry in the diagonal has a value of $\step (\boldsymbol{z}_A^{(a(l)+1)})$. Because the actual sign of neuron $i$ at layer $a(l)+1$ varies by the input, we provide a simple estimation. Given that the neuron $i \in \natset{p_A^{(a(l)+1)}}$ has a pre-activation vector over t samples as $\boldsymbol{z}_{\boldsymbol{.},i} \in \RR^{t}$, which is the $i^{th}$ column of the pre-activation matrix $\boldsymbol{Z}_A^{(a(l)+1)}$. We estimate the $i^{th}$ entry in the diagonal of matrix $\boldsymbol{D}_A^{(a(l)+1)}$ using either sign of sum (i.e., $\step(\sum_{j=1}^t \boldsymbol{z}_{j,i})$) or majority vote (i.e., 1 if at least $\frac{t}{2}$ samples have positive activation values and 0 otherwise) or average of sign ($\frac{1}{t} \sum_{j=1}^t \text{step}(\boldsymbol{z}_{j,i})$).

\vspace{0.5em}\noindent
\textbf{Toy example.} We illustrate two layer balancing methods by a toy example in Figure~\ref{fig:layer_balancing}. We used two neural networks with the same setting as in Figure~\ref{fig:cross_layer_alignment}. Based on the optimal CLA, we can either add layers into Model B or merge layers in Model A to equalize the number of layers.

\vspace{0.5em}\noindent
\textbf{Pros and cons of balance methods.} An advantage of merging layers is that the fused model has fewer layers. However, merging layers degrades the accuracy of model A and it is slower than adding layers method that does not involve any complex calculations. On the flip side, adding layers does not affect the accuracy of model B but results in a deeper fused model. Surprisingly, merging layers shows comparatively better performance than adding layers in our experiments on the multilayer perceptron (MLP). Performance comparison between the two methods will be provided in Appendix~\ref{sec:ablation_studies} and~\ref{subsec:addexp_skill_transfer}. Note that we can use more sophisticated model compression methods instead of the merging layers method. Similarly, it is possible to replace the adding layers method with a more advanced network expanding technique~\cite{wei2016network}. Here, we just introduce two natural and fast ways to demonstrate our framework.

\vspace{0.5em}\noindent
\textbf{Balancing the number of layers for CNN.} The same merging method does not work in the case of CNN. On the other hand, the adding layers method can be easily applied for convolution layers. For VGG, we can set all filters of a new convolution layer to identity kernels. For RESNET, we add a new block in which all filters of two convolution layers become zero kernels while the short-cut connection remains as the identity mapping.

\subsection{Extension to multiple neural networks}
\label{subsec:multiple_networks}
Solving the CLA problem for multiple neural networks is a non-trivial task. It can be formulated as a multi-index assignment problem~\cite{spieksma2000multi,huang2003hybrid}. In addition, it has some connections to the multi-marginal optimal partial optimal transport~\cite{figalli2010optimal,kitagawa2015multi}. Therefore, it is an interesting direction for our future work. Here we only propose a simple yet effective approach to extend CLAFusion to the case of multiple networks. Consider K pre-trained models $\{ M_i \}_{k=1}^K$. Our approach is similar to what they did in OTFusion~\cite[Section~4]{singh2020model}. Starting with an estimation of the fused model $M_{\Ff}$, we apply the first and second parts of CLAFusion $K$ times to depth-align $K$ pre-trained models with respect to the fused model. After that, we apply OTFusion to those depth-aligned networks to produce the final weights for the fused model. The choice of $M_{\Ff}$ plays an important role in this approach. However, it is unclear how it is chosen in OTFusion. In our experiments, the network with the most number of layers is chosen as the initialization for $M_{\Ff}$.

\section{Experiments}
\label{sec:experiments}
\textbf{Outline.} In this section, we showcase the practical usage of CLAFusion in different setups. We first fuse a RESNET34 and a RESNET18 trained on CIFAR10, CIFAR100 ~\cite{krizhevsky2009learning} and Tiny-ImageNet datasets~\cite{chrabaszcz2017downsampled}. Secondly, we incorporate CLA into the framework of heterogeneous model transfer to boost the performance. Finally, our method is applied to the teacher-student setting in which VGG architectures are utilized. In all experiments, OTFusion is adopted as the layer-wise model fusion in the third step of our framework. The detailed settings including training hyperparameters, used assets, and computational resources are deferred to Appendix~\ref{sec:experiment_settings}. We further conduct ablation studies to validate the effectiveness of cross-layer alignment in all three types of architectures (MLP, CNN, and RESNET) and compare two layer balancing methods in Appendix~\ref{sec:ablation_studies}. In Appendix~\ref{sec:addexp}, we provide additional experimental results including skill transfer on the synthetic dataset and supplementary results for experiments in the main text as well as highlight the computational advantage of CLAFusion.

\vspace{0.5em}\noindent
\textbf{Comparison with NeuralMerger.} Though NeuralMerger also considers the heterogeneous model fusion setting, its settings are quite different from ours. NeuralMerger combined VGG-16 with ZFNet~\cite{zeiler2014visualizing} and LeNet~\cite{lecun1998gradient} using hand-designed cross-layer alignment. While our work only merged models of the same architecture family (e.g., VGG-11 and VGG-13). In addition, its code is not available to reproduce the methods in the paper. Furthermore, we also tried to compare our optimal CLA mapping with its mapping by adopting its mapping into our framework while keeping the second and third parts. However, OTFusion, the third component used in our experiments, cannot combine with its mapping. This is because OTFusion requires the height and width of output images (activation matrix) at two corresponding layers of two models to be the same to calculate the cost matrix if using activation-based alignment. If using weight-based alignment, OTFusion requires kernels at two corresponding layers of two models to have the same size. Neither of these two requirements is fulfilled by the hand-crafted CLA in NeuralMerger. Note that NeuralMerger is able to combine weights of two models because it used vector quantization to decompose weights of two models into the same lower dimension while OTFusion is based on the weight averaging mechanism.

\begin{table*}[t!]
    \centering
    \caption{The results of fusing and finetuning RESNET34 ($M_A$) and RESNET18 ($M_B$) on CIFAR10, CIFAR100, and Tiny-ImageNet datasets. Each entity shows the average classification accuracy over 5 seeds. The best performing initialization in each row is in bold. The detailed results are reported in Table~\ref{table:fuse_two_models_details}.}
    \scalebox{0.75}{
        \begin{tabu}{cccccccccc}
        \toprule
        Dataset & $M_A$ & $M_B$ & Ensemble & $M_{\Ff}$ & \multicolumn{5}{c}{Finetune} \\
        \cmidrule(lr){6-10}
        & & & Learning & & $M_A$ & $M_B$ & $M_{\Ff}$ & $M_B$ depth-aligned & Random RESNET34 \\
        \midrule
        CIFAR10 & 93.31 & 92.92 & 93.81 & 65.72 & 93.52 & 93.29 & \textbf{93.67} & 93.24 & 92.00 \\
        CIFAR100 & 65.93 & 65.33 & 68.49 & 27.93 & 66.87 & 66.19 & \textbf{67.59} & 66.48 & 64.59 \\
        Tiny-ImageNet & 54.84 & 53.29 & 58.70 & 17.19 & 55.17 & 53.81 & \textbf{55.46} & 53.94 & 54.16 \\
        \bottomrule
        \end{tabu}
    }
    \label{table:fuse_two_models}
\end{table*}

\subsection{Generate an efficient initialization}
\label{subsec:eff_init}
\textbf{Settings.} Following the experiments conducted in ~\cite[Section~5.3]{singh2020model}, CLAFusion can be used to generate an initialization when training the larger neural network. We apply CLAFusion to pre-trained RESNET34 and RESNET18 on CIFAR10, CIFAR100, and Tiny-ImageNet datasets. Since our purpose is not to obtain state-of-the-art performance, we use the same architecture for both datasets. Layer representation is the pre-activation matrix of $200$ samples while the cost function is calculated by subtracting the linear CKA from 1. After fusing, we also perform finetuning to improve the accuracy as observed in previous model fusion works~\cite{chou2018unifying,singh2020model}. The finetuning hyperparameters, which are adopted from~\cite[Appendix~S3.1.3]{singh2020model}, are reported in Table~\ref{table:finetune_resnet_settings}.

\vspace{0.5em}\noindent
\textbf{Baselines.} We use NET2NET to generate a RESNET34 model, referred to as $M_B$ depth-aligned, which preserves the performance of $M_B$. To compare with our method, we finetune the pre-trained models, $M_B$ depth-aligned, and a RESNET34 from scratch. As a reference, we report the result of ensemble learning that calculates the average predictions over all individual models.

\vspace{0.5em}\noindent
\textbf{Results.} As can be seen from Table~\ref{table:fuse_two_models} retraining from the fused model gains accuracies of $93.67, 67.59$, and $55.46$, which are the highest among all initializations on each dataset. This demonstrates the advantage of CLAFusion over NET2NET when initializing a large model from a pre-trained small model. Because our method allows combining the knowledge from the pre-trained large model to generate a better initialization, rather than solely relying on the knowledge of the small model. Although ensemble learning achieves higher performance, it comes with the cost of a nearly $50\%$ increase in both memory and inference time according to Table~\ref{table:inference_time}.

\begin{table*}[t!]
    \centering
    \caption{The results of fusing and finetuning more than 2 residual networks on CIFAR10, CIFAR100, and Tiny-ImageNet datasets. Pre-trained models are alternately RESNET34 and RESNET18. Each entity shows the average classification accuracy over 5 seeds. The best performing initialization in each row is in bold. The detailed results are reported in Tables~\ref{table:fuse_four_models_details} and~\ref{table:fuse_six_models_details}.}
    \scalebox{0.75}{
        \begin{tabu}{clcclc}
        \toprule
        Dataset & Pre-trained models & Ensemble & $M_{\Ff}$ & \multicolumn{2}{c}{Finetune} \\
        \cmidrule(lr){5-6}
        & & Learning & & Pre-trained models & $M_{\Ff}$ \\
        \midrule
        CIFAR10 & 93.31, 92.92, 93.16, 92.83 & 94.17 & 31.76 & 93.52, 93.29, 93.41, 93.20 & \textbf{93.76} \\
        & 93.31, 92.92, 93.16, 92.83, 93.18, 92.86 & 94.22 & 23.75 & 93.52, 93.29, 93.41, 93.20, 93.49, 93.24 & \textbf{93.78} \\
        \midrule
        CIFAR100 & 65.93, 65.33, 65.97, 65.28 & 70.27 & 8.47 & 66.87, 66.19, 66.77, 65.96 & \textbf{68.92} \\
        & 65.93, 65.33, 65.97, 65.28, 65.89, 65.30 & 70.62 & 4.39 & 66.87, 66.19, 66.77, 65.96, 66.56, 65.92 & \textbf{69.40} \\
        \midrule
        Tiny-ImageNet & 54.84, 53.29, 54.91, 53.23 & 61.28 & 4.05 & 55.17, 53.81, 55.29, 53.76 & \textbf{56.07} \\
        & 54.84, 53.29, 54.91, 53.23, 54.63, 53.47 & 62.32 & 2.47 & 55.17, 53.81, 55.29, 53.76, 55.06, 54.04 & \textbf{56.47} \\
        \bottomrule
        \end{tabu}
    }
    \label{table:fuse_multiple_models}
\end{table*}

\vspace{0.5em}\noindent
\textbf{More than 2 models.} We consider the same settings as in the 2-model scenario but fuse 4(6) models instead. We train 2(3) RESNET34 and 2(3) RESNET18 models and fuse them using CLAFusion before finetuning. In Table~\ref{table:fuse_multiple_models}, the first, third (and fifth) pre-trained models are RESNET34 while the second, fourth (and sixth) pre-trained models are RESNET18. The fused model, after finetuning, consistently yields the best performance on both datasets. On CIFAR100, for example, the classification accuracy increases remarkably over finetuning pre-train models by a margin of at least $2.05$ and $2.53$ for the cases of 4 and 6 models, respectively. In addition, increasing the number of pre-trained models increases the final performance of CLAFusion. Our hypothesis is that fusing multiple models which were trained at different random seeds can aggregate different “knowledge” from these models. This is similar to the outcome of ensemble learning. In terms of computational resources, the gap between our model fusion method and ensemble learning is pushed even further. Table~\ref{table:inference_time} shows that ensemble learning uses approximately 3 and 4.5 times as many resources as CLAFusion when combining 4 and 6 pre-trained RESNET models, respectively.

\begin{table*}[t!]
    \centering
    \caption{The results of heterogeneous model transfer from RESNET18 to RESNET34. Each entity shows the average classification accuracy $\pm$ standard deviation over 5 seeds. The best performing method in each row is in bold. The detailed results can be found in Table~\ref{table:model_transfer_details}.}
    \scalebox{0.75}{
        \begin{tabular}{cccccc}
        \toprule
        Dataset & Method & HMT & HMT + Naive avg & HMT + OTFusion & HMT + CLA + OTFusion \\
        \midrule
        CIFAR10 & Transfer & 10.79 $\pm$ 1.01 & 37.49 $\pm$ 4.59 & 18.32 $\pm$ 1.76 & \textbf{61.29 $\pm$ 3.69} \\
        & Transfer + Finetune & 93.10 $\pm$ 0.18 & 93.38 $\pm$ 0.19 & 93.66 $\pm$ 0.14 & \textbf{93.86 $\pm$ 0.14} \\
        \midrule
        CIFAR100 & Transfer & 2.70 $\pm$ 0.45 & 10.79 $\pm$ 1.36 & 5.14 $\pm$ 1.22 & \textbf{12.56 $\pm$ 2.79} \\
        & Transfer + Finetune & 65.15 $\pm$ 0.24 & 66.42 $\pm$ 0.25 & 67.23 $\pm$ 0.16 & \textbf{67.83 $\pm$ 0.21} \\
        \bottomrule
        \end{tabular}
    }
    \label{table:model_transfer}
    \vskip 0.1in
\end{table*}

\subsubsection{Heterogeneous model transfer}
\label{subsubsec:hmt}
\textbf{Settings.} Heterogeneous model transfer~\cite{wang2021hmt} is a branch of model transfer that deals with heterogeneous neural networks. It is contrasted to homogeneous model transfer, also known as transfer learning, in which the same network architecture is used in both the pre-training and finetuning phase. In this experiment, we apply CLA and OTFusion to the heterogeneous model transfer method (in short, HMT). We transfer the pre-trained RESNET18 to the pre-trained RESNET34 using different methods, then finetune and compare their performance. We use the same pre-trained models, layer representation, layer dissimilarity, and finetuning hyperparameters as in Section~\ref{subsec:eff_init}.

\vspace{0.5em}\noindent
\textbf{Baselines.} First, the heterogeneous model transfer method (HMT) in the original paper~\cite{wang2021hmt} is used. Model parameters of the pre-trained RESNET18 are transferred to the pre-trained RESNET34. The weights that are not transferred are set to the weights of the pre-trained RESNET34. In the second method (HMT + Naive avg), after transferring we apply naive averaging to the transferred RESNET34, which is the result of HMT, and the pre-trained RESNET34. Naive averaging is replaced with OTFusion in the third method (HMT + OTFusion). Different from the third method, the last method utilizes the optimal mapping from the CLA problem instead of the longest chain proposed in Wang et al.~\cite{wang2021hmt} for the layer-to-layer transfer.

\vspace{0.5em}\noindent
\textbf{Results.} Table~\ref{table:model_transfer} demonstrates the average performance of different transfer methods. Using CLA improves the performance over the longest chain substantially, with a significant gap of $42.97$ and $7.42$ between the third and fourth methods on CIFAR10 and CIFAR100 datasets, respectively. After finetuning, the combination of CLA and OTFusion leads to an improvement of $0.76$ and $2.68$ compared to using HMT only on CIFAR10 and CIFAR100 datasets, respectively. In addition, its final performance on CIFAR10 and CIFAR100 (93.86 and 67.83) is also better than finetuning pre-trained models in Table~\ref{table:fuse_two_models} (93.52 and 66.87) while that of HMT does not.

\begin{table*}[t!]
    \centering
    \caption{The results of fusing teacher and student VGG models on the CIFAR10 dataset. The best performing student model is in bold. More results can be found in Table~\ref{table:teacher_student_details}.}
    \scalebox{0.9}{
    \begin{tabular}{cccccccc}
        \toprule
        \# Params & Teacher & \multicolumn{2}{c}{Students} & \multicolumn{4}{c}{Finetune} \\
        \cmidrule(lr){3-4} \cmidrule(lr){5-8}
        $(M_A, M_B, M_{\Ff})$ & $M_A$ & $M_B$ & $M_{\Ff}$ & $M_A$ & $M_B$ & $M_{\Ff}$ & $M_B$ depth-aligned  \\
        \midrule
        (33M, 3M, 3M) & 92.70 & 89.92 & 82.66 & 92.65 & 89.89 & \textbf{90.96} & 90.84 \\
        \bottomrule
    \end{tabular}
    }
    \label{table:teacher_student_fusion}
    \vskip 0.1in
\end{table*}

\begin{table*}[t!]
    \centering
    \caption{Knowledge distillation from teacher $M_A$ into student model $M_B$. The "Best`` row shows the best performance for all combinations of hyperparameters. The "Average" row illustrates the mean $\pm$ standard deviation of the best accuracies obtained at different temperatures. The best performing initialization in each row is in bold. The detailed results are reported in Table~\ref{table:knowledge_distillation_details}.}
    \scalebox{0.9}{
    \begin{tabular}{cccccc}
        \toprule
        & \multicolumn{5}{c}{Distillation initialization} \\
        \cmidrule(lr){2-6}
        & Random $M_B$ & $M_B$ & $M_{\Ff}$ & $M_B$ depth-aligned & Random $M_{\Ff}$ \\
        \midrule
        Best & 89.10 & 90.98 & \textbf{91.43} & 91.24 & 88.84
        \\
        Average & 88.41 $\pm$ 0.77 & 90.73 $\pm$ 0.43 & \textbf{91.16 $\pm$ 0.25} & 91.09 $\pm$ 0.19 & 88.10 $\pm$ 0.68
        \\
        \bottomrule
    \end{tabular}
    }
    \label{table:knowledge_distillation}
\end{table*}

\subsection{Teacher-student fusion}
\label{subsec:teacher_student_fusion}
\textbf{Settings.} In this experiment, we transfer the knowledge from the pre-trained teacher model to the student model. We train two VGG models on the CIFAR10 dataset: model A has VGG13 architecture with a double number of channels at each layer while model B has VGG11 architecture with a half number of channels at each layer. After fusing the teacher and student models, we retrain the fused model and compare it to the retraining result of the student model. The layer measure and the layer dissimilarity are identical to Section~\ref{subsec:eff_init}. We adopt the best hyperparameters as reported in~\cite[Appendix~S11]{singh2020model} for retraining. All finetuning hyperparameters are reported in Table~\ref{table:finetune_vgg_settings}. Additional experiments with teacher and student RESNET models on CIFAR100 are given in Appendix~\ref{subsec:addexp_teacher_student_fusion}.

\vspace{0.5em}\noindent
\textbf{Results.} The classification accuracies are summarized in Table~\ref{table:teacher_student_fusion}. After retraining, the fused model yields better performance than retraining other student models. The compression ratio between the fused model and the teacher is about 11 at a cost of reducing $1.74\%$ accuracy. This suggests that CLAFusion can act as a network pruning method to compress a heavy model into a lightweight one.

\subsubsection{Knowledge distillation}
\label{subsubsec:knowledge_distillation}
\textbf{Settings.} It has been shown that pre-trained distillation~\cite{turc2019well}, which performs knowledge distillation from pre-trained large teacher models to a pre-trained student model, outperforms both knowledge distillation to a randomly initialized student and the conventional pre-training + finetuning scheme. As discussed in OTFusion, the fused model can be used as an efficient initialization for knowledge distillation. To examine the same application of CLAFusion, we perform knowledge distillation to the above pre-trained VGG models. We employ the method in Hinton et al.~\cite{hinton2015distilling} to match the logit distribution of the student model to that of the teacher model. For initialization, we consider five different choices for the student model: (a) randomly initialized $M_B$, (b) $M_B$, (c) $M_{\Ff}$, (d) $M_B$ depth-aligned, (e) randomly initialized $M_{\Ff}$. We sweep over a set of hyperparameters to choose the best combination that maximizes the accuracy of the student model. The sets of hyperparameters for distillation also follows the setting in~\cite[Appendix~S12]{singh2020model}: temperature $T = \{20, 10, 8, 4, 1\}$ and loss-weight factor $\gamma = \{ 0.05, 0.1, 0.5, 0.7, 0.95, 0.99 \}$. 

\vspace{0.5em}\noindent
\textbf{Results.} The results for knowledge distillation are reported in Table~\ref{table:knowledge_distillation}. Using the fused model as an initialization achieves the best performance among different choices of initializations. It showcases the application of CLAFusion as an efficient initialization for knowledge distillation. If averaging over different temperatures, retraining the fused model $(90.96)$ works better than pre-trained distillation $(90.73)$. Even the best accuracy of pre-trained distillation into the pre-trained student model $(90.98)$ is slightly higher than that of retraining the fused model but comes with the cost of hyperparameter tuning. Moreover, as reported in Table~\ref{table:training_time}, an epoch in pre-trained distillation takes 12.22 seconds while that of finetuning only costs 7.48 seconds, which is around 1.63x speedup. It further strengthens the claim that CLAFusion can serve as a reliable model compression method.

\section{Conclusion}
\label{sec:conclusion}
In the paper, we extend layer-wise model fusion methods to the setting of heterogeneous neural networks by solving a cross-layer alignment problem, followed by a layer balancing step. Finetuning the fused network from CLAFusion achieves a better accuracy for RESNET trained on the CIFAR10, CIFAR100, and Tiny-ImageNet datasets. In addition, our framework can be incorporated into the heterogeneous model transfer framework to improve performance. Furthermore, it shows potential applications for model compression and knowledge distillation.



\appendix
\begin{center}
\textbf{\Large{Supplement to ``Model Fusion of Heterogeneous Neural Networks via Cross-Layer Alignment''}}
\end{center}

\noindent
In this supplementary material, we first present the limitations of our method; then, detail the common settings, used assets, and computational resources in Appendix~\ref{sec:experiment_settings}. Next, we illustrate the efficiency of CLA as well as the importance of constraints on the first and last layers in Appendix~\ref{sec:ablation_studies}. Finally, we provide additional results of our experiments in Appendix~\ref{sec:addexp}.

\section{Limitations}
The major limitations of our framework stem from our constraints on the architecture of neural networks. Because in all experiments, OTFusion is used in the third step of our framework, CLAFusion shares the same limitations as OTFusion. Firstly, we do not consider networks that include bias and batch normalization though they can be handled in the same way as discussed in OTFusion~\cite[Appendix~S1]{singh2020model}. Secondly, our framework is not generalized to complex architectures such as LSTM~\cite{hochreiter1997long} or Transformer~\cite{vaswani2017attention}. The main reason is due to a lack of an effective approach for combining weights of complex components such as LSTM cell or attention module in a reasonable way. This is a common limitation for all model fusion methods based on weight averaging mechanism~\cite{utans1996weight,smith2017investigation,singh2020model}. Thirdly, while CLAFusion can combine two networks of different depths and different widths, two networks must belong to the same architecture family. This is because OTFusion requires the height and width of output images (activation matrix) at two corresponding layers of two models to be the same to calculate the cost matrix if using activation-based alignment. If using weight-based alignment, OTFusion requires kernels at two corresponding layers of two models to have the same size. Fourthly, similar to other model fusion approaches~\cite{utans1996weight,smith2017investigation,Leontev_2019,wang2020federated,singh2020model}, an additional fine-tuning process is needed after performing model fusion to help the fused model surpass the performance of the original pre-trained models. In terms of layer-balancing techniques, merging layers degrades the performance of original models and incurs computational overhead. In addition, our current merging technique is not applicable to convolutional layers.

\section{Experiment settings}
\label{sec:experiment_settings}
In all experiments, we use OTFusion to fuse same-length models after aligning and balancing in the first two steps of CLAFusion. Following the same alignment strategy in~\cite[Appendix~S1]{singh2020model}, OTFusion is computed using the activation-based alignment strategy in which the pre-activation matrix instead of the activation matrix is used for the neuron measure. The hyperparameters for pre-trained models are summarized in Table~\ref{table:training_details}. Without any further specification, the random seed for training or retraining is set to its default value. The accuracy of pre-trained MLP is the accuracy of the last epoch. While the accuracies of VGG and RESNET are reported as the best performing checkpoint. For all finetuning experiments, we always chose the best record among all epochs. The hyperparameters for finetuning RESNET and VGG models are reported in Tables~\ref{table:finetune_resnet_settings} and~\ref{table:finetune_vgg_settings}, respectively.

\begin{table}[t!]
    \centering
    \caption{The hyperparameters for training different network architectures.}
    \scalebox{0.95}{
    \begin{tabular}{lcccc}
        \toprule
        & MLP & VGG & ResNet (CIFAR) & RESNET (Tiny-ImageNet) \\
        \midrule
        Number of epochs & 10 & 300 & 300 & 120 \\
        Training batch size & 64 & 128 & 256 & 256 \\
        Test batch size & 1000 & 1000 & 1000 & 1000 \\
        Optimizer & SGD & SGD & SGD & SGD \\
        Initial LR & 0.01 & 0.05 & 0.1 & 0.1 \\
        Momentum & 0.5 & 0.9 & 0.9 & 0.9 \\
        Weight decay & & 0.0005 & 0.0001 & 0.0001 \\
        LR decay factor & & 2 & 10 & 10 \\
        LR decay epochs & & 30,60,$\ldots$,270 & 150,250 & 30,60,90 \\
        Default seed & 0 & 42 & 42 & 42 \\
        5 seeds & 0,1,2,3,4 & 40,41,42,43,44 & 40,41,42,43,44 & 40,41,42,43,44 \\
        \bottomrule
    \end{tabular}
    }
    \label{table:training_details}
\end{table}

\begin{table}[t!]
    \parbox{.5\linewidth}{
        \centering
        \caption{Finetuning RESNET hyperparameters.}
        \scalebox{0.9}{
            \begin{tabular}{lcc}
            \toprule
            & CIFAR & Tiny-ImageNet \\
            \midrule
            Number of epochs & 120 & 90 \\
            Training batch size & 256 & 256 \\
            Test batch size & 1000 & 1000 \\
            Optimizer & SGD & SGD \\
            Initial LR & 0.1 & 0.1 or 0.01\\
            Momentum & 0.9 & 0.9 \\
            Weight decay & 0.0001 & 0.0001 \\
            LR decay factor & 2 & 10 \\
            LR decay epochs & 20,40,60,80,100 & 30,60,90 \\
            \bottomrule
            \end{tabular}
        }
        \label{table:finetune_resnet_settings}
    }
    \hfill
    \parbox{.45\linewidth}{
        \centering
        \caption{Finetuning VGG hyperparameters.}
        \scalebox{0.9}{
            \begin{tabular}{lcc}
            \toprule
            & CIFAR \\
            \midrule
            Number of epochs & 120 \\
            Training batch size & 128 \\
            Test batch size & 1000 \\
            Optimizer & SGD \\
            Initial LR & 0.01 \\
            Momentum & 0.9\\
            Weight decay & 0.0005 \\
            LR decay factor & 2 &  \\
            LR decay epochs & 20,40,60,80,100 \\
            \bottomrule
            \end{tabular}
        }
        \label{table:finetune_vgg_settings}
    }
\end{table}

\vspace{0.5em}\noindent
\textbf{Open source code.} We adapt the official implementation of OTFusion\footnote{\href{https://github.com/sidak/otfusion}{https://github.com/sidak/otfusion}}~\cite{singh2020model} in our implementation. For computing the optimal transport, we use Python Optimal Transport (POT) library\footnote{\href{https://github.com/PythonOT/POT}{https://github.com/PythonOT/POT}}~\cite{flamary2021pot}. For model transfer, the source code of the original paper\footnote{\href{https://anonymous.4open.science/r/6ab184dc-3c64-4fdd-ba6d-1e5097623dfd/a\_hetero\_model\_transfer.py}{https://anonymous.4open.science/r/6ab184dc-3c64-4fdd-ba6d-1e5097623dfd/a\_hetero\_model\_transfer.py}}~\cite{wang2021hmt} is utilized.

\vspace{0.5em}\noindent
\textbf{Neural network architectures.} Abusing the notation in VGG's paper~\cite{simonyan2014very}, we represent several CNN architectures used in the experiments as follows. 

\noindent \emph{VGG13 doub}: Input layer $\rightarrow \text{conv3-64} \rightarrow \text{conv3-128} \rightarrow \text{maxpool} \rightarrow \text{conv3-256} \rightarrow \text{conv3-256} \rightarrow \text{maxpool} \rightarrow \text{conv3-512} \rightarrow \text{conv3-512} \rightarrow \text{maxpool} \rightarrow \text{conv3-1028} \rightarrow \text{conv3-1028} \rightarrow \text{maxpool} \rightarrow \text{conv3-1028} \rightarrow \text{conv3-512} \rightarrow \text{maxpool} \rightarrow $ Output layer

\vspace{0.5em}\noindent
\emph{VGG11 half}: Input layer $\rightarrow \text{conv3-64} \rightarrow \text{maxpool} \rightarrow \text{conv3-64} \rightarrow \text{maxpool} \rightarrow \text{conv3-128} \rightarrow \text{conv3-128} \rightarrow \text{maxpool} \rightarrow \text{conv3-256} \rightarrow \text{conv3-256} \rightarrow \text{maxpool} \rightarrow \text{conv3-256} \rightarrow \text{conv3-512} \rightarrow \text{maxpool} 
\rightarrow $ Output layer

\vspace{0.5em}\noindent
\emph{VGG13 stud}: Input layer $\rightarrow \text{conv3-64} \rightarrow \text{conv3-64} \rightarrow \text{maxpool} \rightarrow \text{conv3-64} \rightarrow \text{conv3-64} \rightarrow \text{maxpool} \rightarrow \text{conv3-128} \rightarrow \text{conv3-128} \rightarrow \text{maxpool} \rightarrow \text{conv3-256} \rightarrow \text{conv3-256} \rightarrow \text{maxpool} \rightarrow \text{conv3-256} \rightarrow \text{conv3-512} \rightarrow \text{maxpool} \rightarrow $ Output layer

\vspace{0.5em}\noindent
\emph{RESNET34}: Input layer $
\rightarrow \text{conv3-64} \rightarrow \text{conv3-64} \rightarrow \text{conv3-64} \rightarrow \text{conv3-64} \rightarrow \text{conv3-64} \rightarrow \text{conv3-64} \rightarrow \text{conv3-64}
\rightarrow \text{conv3-128} \rightarrow \text{conv3-128} \rightarrow \textcolor{red}{conv1-128} \rightarrow \text{conv3-128} \rightarrow \text{conv3-128} \rightarrow \text{conv3-128} \rightarrow \text{conv3-128} \rightarrow \text{conv3-128} \rightarrow \text{conv3-128} 
\rightarrow \text{conv3-256} \rightarrow \text{conv3-256} \rightarrow \textcolor{red}{conv1-256} \rightarrow \text{conv3-256} \rightarrow \text{conv3-256} \rightarrow \text{conv3-256} \rightarrow \text{conv3-256} \rightarrow \text{conv3-256} \rightarrow \text{conv3-256} \rightarrow \text{conv3-256} \rightarrow \text{conv3-256} \rightarrow \text{conv3-256} \rightarrow \text{conv3-256} 
\rightarrow \text{conv3-512} \rightarrow \text{conv3-512} \rightarrow \textcolor{red}{conv1-512} \rightarrow \text{conv3-512} \rightarrow \text{conv3-512} \rightarrow \text{conv3-512} \rightarrow \text{conv3-512} \rightarrow$ Output layer

\vspace{0.5em}\noindent
\emph{RESNET18}: Input layer $
\rightarrow \text{conv3-64} \rightarrow \text{conv3-64} \rightarrow \text{conv3-64} \rightarrow \text{conv3-64} \rightarrow \text{conv3-64} 
\rightarrow \text{conv3-128} \rightarrow \text{conv3-128} \rightarrow \textcolor{red}{conv1-128} \rightarrow \text{conv3-128} \rightarrow \text{conv3-128} 
\rightarrow \text{conv3-256} \rightarrow \text{conv3-256} \rightarrow \textcolor{red}{conv1-256} \rightarrow \text{conv3-256} \rightarrow \text{conv3-256} 
\rightarrow \text{conv3-512} \rightarrow \text{conv3-512} \rightarrow \textcolor{red}{conv1-512} \rightarrow \text{conv3-512} \rightarrow \text{conv3-512} \rightarrow$ Output layer

\vspace{0.5em}\noindent
Note that the average-pooling layer is skipped in the representation. In addition, the last fully-connected layers, which are usually called classifier layers, are merged into the output layer. In RESNET models, convolution layers in red color with a kernel size of 1 is the shortcut layer that connects two stages of different channel sizes. These layers are represented by dotted shortcuts in Figure 3 in RESNET's paper~\cite{he2016deep}.

\vspace{0.5em}\noindent
\textbf{Computing devices.} All experiments are done on 1 NVIDIA A100 40GB GPU. 

\section{Ablation studies}
\label{sec:ablation_studies}
In this section, we show that the optimal CLA is superior to other strictly increasing mappings by comparing the performance of the fused model which is obtained by applying the second and third steps of CLAFusion with the corresponding cross-layer mapping. We first study the importance of CLA and two constraints to the classification accuracy of the fused model when the pre-trained models are fully connected neural networks. In the latter part, we validate the advantage of CLA in fusing convolutional neural networks.

\begin{table}[t!]
    \centering
    \caption{The optimal CLA for different combinations of layer representation and cost function when fusing 2 MLPs on the MNIST dataset. Experiments were run with 5 different seeds. Changing the random seed did not alter the optimal mappings except for two cases which are in red.}
    \scalebox{0.84}{
    \begin{tabular}{llcccc}
        \toprule
        Layer representation & Cost function & Both & Last layer constraint & First layer constraint & None \\
        \midrule
        Number of neurons & squared difference & [1, 2, 5] & [1, 2, 5] & [1, 2, 3] & [1, 2, 3] \\
        Pre-activation matrix & 1 - linear CKA & [1, 2, 5] & [1, 2, 5] & \textcolor{red}{[1, 2, 3]}  & \textcolor{red}{[1, 2, 3]} \\
        Pre-activation matrix & Wasserstein distance & [1, 2, 5] & [1, 2, 5] & [1, 2, 3] & [1, 2, 3] \\
        \bottomrule
    \end{tabular}
    }
    \label{table:layer_alignment_results}
    \vskip 0.1in
\end{table}

\subsection{Fully connected neural networks}
\textbf{Settings.} We compare the performance of different combinations of layer representation and layer balancing methods in our framework. To prove the efficiency of the CLA step, we compare the result of fusing two models using different mappings. We also consider removing the constraint on the first and last hidden layers. We train two MLPs on the MNIST dataset~\cite{lecun1998gradient} in 5 different seeds. Model A has 5 hidden layers of size 400, 200, 100, 50, 25 while model B has 3 hidden layers of size 400, 200, 100 as the following representation. 

\vspace{0.5em}\noindent
\emph{Model A}: Input layer $\rightarrow \underbrace{\text{FC-400}}_\text{layer 1} \rightarrow \underbrace{\text{FC-200}}_\text{layer 2} \rightarrow \underbrace{\text{FC-100}}_\text{layer 3} \rightarrow \underbrace{\text{FC-50}}_\text{layer 4} \rightarrow \underbrace{\text{FC-25}}_\text{layer 5} \rightarrow$ Output layer

\vspace{0.5em}\noindent
\emph{Model B}: Input layer $\rightarrow \underbrace{\text{FC-400}}_\text{layer 1} \rightarrow \underbrace{\text{FC-200}}_\text{layer 2} \rightarrow \underbrace{\text{FC-100}}_\text{layer 3} \rightarrow$ Output layer

\begin{table}[t!]
    \centering
    \caption{Performance comparison between different combinations of mapping and balancing methods when fusing 2 MLPs on the MNIST dataset. Experiments were run with 5 different seeds. The mapping obtained from the CLA step and the best performing mapping in  columns $M_{\Ff}$ are in bold.}
    \scalebox{1.0}{
    \begin{tabular}{ccccccc}
        \toprule
        Mapping & $M_A$ & $M_B$ & \multicolumn{3}{c}{$M_{\Ff}$} \\
        \cmidrule(lr){4-6}
        & & & Add & Merge (sum) & Merge (average) \\
        \midrule
        $[1, 2, 3]$ & \multirow{10}{*}{96.95 $\pm$ 0.18} & \multirow{10}{*}{97.75 $\pm$ 0.04} & 92.33 $\pm$ 2.35 & 93.90 $\pm$ 1.57 & 91.84 $\pm$ 3.06 \\
        $[1, 2, 4]$ & & & 92.48 $\pm$ 2.07 & 93.41 $\pm$ 1.56 & 91.88 $\pm$ 1.62 \\
        $\boldsymbol{[1, 2, 5]}$ & & & \textbf{92.49 $\pm$ 2.64} & \textbf{94.00 $\pm$ 1.36} & \textbf{92.37 $\pm$ 2.94} \\
        $[1, 3, 4]$ & & & 90.69 $\pm$ 2.62 & 92.21 $\pm$ 1.86 & 89.17 $\pm$ 0.98 \\
        $[1, 3, 5]$ & & & 90.88 $\pm$ 2.61 & 93.04 $\pm$ 2.44 & 89.57 $\pm$ 3.84 \\
        $[1, 4, 5]$ & & & 90.98 $\pm$ 2.35 & 93.06 $\pm$ 2.07 & 90.26 $\pm$ 2.84 \\
        $[2, 3, 4]$ & & & 82.21 $\pm$ 1.79 & 51.26 $\pm$ 6.41 & 55.62 $\pm$ 5.97 \\
        $[2, 3, 5]$ & & & 83.14 $\pm$ 2.18 & 56.56 $\pm$ 5.87 & 59.46 $\pm$ 5.73 \\
        $[2, 4, 5]$ & & & 83.00 $\pm$ 2.01 & 54.16 $\pm$ 6.01 & 58.62 $\pm$ 5.60 \\
        $[3, 4, 5]$ & & & 79.49 $\pm$ 2.37 & 56.31 $\pm$ 4.79 & 55.90 $\pm$ 3.02 \\
        \bottomrule
    \end{tabular}
    }
    \label{table:ablation_studies_mlp}
    \vskip 0.1in
\end{table}

\vspace{0.5em}\noindent
\textbf{The optimal CLA.} The cross-layer mappings obtained using Algorithm~\ref{alg:layer_alignment} with different combinations are given in Table~\ref{table:layer_alignment_results}. According to Definition~\ref{def:cross_layer_alignment} in Section~\ref{subsec:cross_layer_alignment}, the optimal mapping is strictly increasing mapping $a : \natset{n} \mapsto \natset{m}$  where $n$ and $m$ are the number of hidden layers of the shallower and deeper networks, respectively. A mapping $a$ is represented as $[a(1), a(2), \ldots, a(n)]$ where $a(i) = j$ indicates that layer i of model B is aligned to layer j of model A. In this case, we have $m = 5$ and $n=3$ and the mapping [1, 2, 5] means that $a(1) = 1, a(2) = 2,$ and $a(3) = 5$. We observe that changing the random seed barely affects the result of CLA in this setting. When keeping both first and last layer constraints, all three combinations yield the same mappings. All three combinations share the same mappings in almost all cases even if none of the two constraints are placed\footnote{When the random seed is 2, the optimal mapping for "pre-activation matrix`` + "1 - linear CKA`` is $[1, 2, 5]$ instead of $[1, 2, 3]$.}. In this case, the first layer constraint does not affect the results of the optimal mapping in all combinations.

\vspace{0.5em}\noindent
\textbf{The effectiveness of the optimal mapping.} We fuse two MLPs for all 10 possible mappings between two models A and B. For the layer balancing method, we try both adding layers and merging layers (the sign of sum and the average of sign estimations). The average performance of pre-trained models and the fused model across 5 random seeds are summarized in Table~\ref{table:ablation_studies_mlp}. On either choice of the layer balancing method, the best accuracy of the fused model is achieved when imposing both the first and last layer constraints. Removing the last layer constraint decreases slightly the performance of the fused model. When the first layers of the two models do not match each other in the last 4 rows, the accuracy drops dramatically. Because an input image may consist of negative cell values due to normalization, adding a new layer as the first hidden layer does not maintain the accuracy of the shallower model. Therefore, it is necessary to match the first and last hidden layers of the two models. Comparing layer balancing methods, the merging layers method runs slower but interestingly leads to a higher accuracy than the adding layers in this situation.

\subsection{Convolutional neural networks}
In this section, we conduct ablation studies on two common CNN architectures including VGG and RESNET. For solving the CLA problem, layer representation is the pre-activation matrix of 200 samples while the cost function is calculated by subtracting the linear CKA from 1. In terms of the layer balancing method, adding layers is chosen. After fusing two models, we perform finetuning with hyperparameters in Tables~\ref{table:finetune_resnet_settings} and~\ref{table:finetune_vgg_settings}.

\subsubsection{VGG architecture}
\textbf{Settings.} We consider the same two VGG models used in Section~\ref{subsec:teacher_student_fusion} on the CIFAR10 dataset. Using the same notation as in VGG's paper~\cite{simonyan2014very}, the architectures of the two models can be represented as follows. 

\vspace{0.5em}\noindent
\emph{Model A}: Input layer $\rightarrow 
\Big[\underbrace{\text{conv3-64}}_\text{layer 1} \rightarrow \underbrace{\text{conv3-128}}_\text{layer 2}\Big] \rightarrow \text{maxpool} \rightarrow 
\Big[\underbrace{\text{conv3-256}}_\text{layer 3} \rightarrow \underbrace{\text{conv3-256}}_\text{layer 4}\Big] \rightarrow \text{maxpool} \rightarrow 
\Big[\underbrace{\text{conv3-512}}_\text{layer 5} \rightarrow \underbrace{\text{conv3-512}}_\text{layer 6}\Big] \rightarrow \text{maxpool} \rightarrow 
\Big[\underbrace{\text{conv3-1028}}_\text{layer 7} \rightarrow \underbrace{\text{conv3-1028}}_\text{layer 8}\Big] \rightarrow \text{maxpool} \rightarrow 
\Big[\underbrace{\text{conv3-1028}}_\text{layer 9} \rightarrow \underbrace{\text{conv3-512}}_\text{layer 10}\Big] \rightarrow \text{maxpool} 
\rightarrow$ Output layer

\vspace{0.5em}\noindent
\emph{Model B}: Input layer $\rightarrow 
\Big[\underbrace{\text{conv3-64}}_\text{layer 1}\Big] \rightarrow \text{maxpool} \rightarrow 
\Big[\underbrace{\text{conv3-64}}_\text{layer 2}\Big] \rightarrow \text{maxpool} \rightarrow 
\Big[\underbrace{\text{conv3-128}}_\text{layer 3} \rightarrow \underbrace{\text{conv3-128}}_\text{layer 4}\Big] \rightarrow \text{maxpool} \rightarrow 
\Big[\underbrace{\text{conv3-256}}_\text{layer 5} \rightarrow \underbrace{\text{conv3-256}}_\text{layer 6}\Big] \rightarrow \text{maxpool} \rightarrow 
\Big[\underbrace{\text{conv3-256}}_\text{layer 7} \rightarrow \underbrace{\text{conv3-512}}_\text{layer 8}\Big] \rightarrow \text{maxpool}
\rightarrow$ Output layer

\vspace{0.5em}\noindent
Note that following the official implementation of OTFusion, we did not modify the number of channels of the first and last convolution layers.

\begin{table}[t!]
    \centering
    \caption{Performance comparison between different mappings when fusing 2 VGG models on the CIFAR10 dataset. Experiments were run with 5 different seeds. To make the notation succinct, the mapping only indicates the values of $a(1)$ and $a(2)$. The mapping obtained from the CLA step is in bold while two additional inter-group mappings are in red. The best performing mapping in columns $M_{\Ff}$ is in bold.}
    \scalebox{1.0}{
    \begin{tabular}{ccccc}
        \toprule
        Mapping & $M_A$ & $M_B$ & \multicolumn{2}{c}{$M_{\Ff}$} \\
        \cmidrule(lr){4-5}
        & & & Before finetuning & After finetuning \\
        \midrule
        \textcolor{red}{$[1, 2]$} & \multirow{6}{*}{92.65 $\pm$ 0.16} & \multirow{6}{*}{89.72 $\pm$ 0.12} & 62.76 $\pm$ 1.92 & 90.56 $\pm$ 0.12 \\
        $\boldsymbol{[1, 3]}$ & & & \textbf{82.03 $\pm$ 0.50} & \textbf{90.64 $\pm$ 0.17} \\
        $[1, 4]$ & & & 79.37 $\pm$ 1.05 & 90.37 $\pm$ 0.20 \\
        $[2, 3]$ & & & 46.64 $\pm$ 2.23 & 89.32 $\pm$ 0.15 \\
        $[2, 4]$ & & & 49.63 $\pm$ 2.80 & 89.03 $\pm$ 0.19 \\
        \textcolor{red}{$[3, 4]$} & & & 46.93 $\pm$ 1.99 & 84.62 $\pm$ 0.16 \\
        \bottomrule
    \end{tabular}
    }
    \label{table:ablation_studies_vgg}
\end{table}

\vspace{0.5em}\noindent
\textbf{The optimal CLA.} Because two models have identical input, output, and fully connected layers, we only need to consider matching their convolution layers. The last layers of the two models are fully connected layers, which are automatically assigned to each other in this setup as we solve the CLA problem for different layer types (e.g., convolutional layers and fully connected layers) independently. As discussed in Section~\ref{subsec:cross_layer_alignment}, each VGG model is divided into 5 groups which are separated by the max-pooling layers and enclosed in square brackets in the above representation. For the last three pairs of groups, the optimal mappings are trivial because each pair has the same number of layers. Abusing the notation in Definition~\ref{def:cross_layer_alignment} in Section~\ref{subsec:cross_layer_alignment}, we have $a(3+i) = 5+i, i \in \natset{5}$. Therefore, we only need to solve the CLA problem for the first two groups, i.e., to find the values of $a(1)$ and $a(2)$. In this case, we again observe that the optimal mapping remains the same even if changing the random seed. 

\vspace{0.5em}\noindent
\textbf{The effectiveness of the optimal mapping.} Other than the optimal mapping obtained from CLA, we fuse two VGG models for 3 intra-group mappings and 2 additional inter-group mappings. Here, intra-group means that layers in the first(second) group of model B are only aligned to layers in the first(second) group of model A. While inter-group means that layers in the first or second group of model B can be aligned to layers in the both first and second groups of model A. In essence, the inter-group mappings are discouraged due to the inconsistency of the input image size as discussed in Section~\ref{subsec:otfusion}. In this experiment only, we consider them for comparison purposes. Table~\ref{table:ablation_studies_vgg} illustrates the average performance of the fused model for different mappings. Before finetuning, it can be seen clearly that the optimal CLA shows a clear margin of at least $2.66\%$ in the performance over other mappings. Similar to the case of MLPs, the performance falls remarkably when not aligning the first layers of two models in the last 3 rows. In terms of the performance after finetuning, the optimal CLA still beats other mappings, resulting in an accuracy of 90.64 on average.

\subsubsection{RESNET architecture}
\textbf{Settings.} We consider the same two RESNET models used in Section~\ref{subsec:eff_init} on the CIFAR10 dataset. Using the same notation as in VGG's paper~\cite{simonyan2014very}, the architectures of the two models can be represented as follows. 

\vspace{0.5em}\noindent
\emph{RESNET34}: Input layer $
\rightarrow \text{conv3-64} \rightarrow \Big[(\text{conv3-64} \rightarrow \text{conv3-64}) \rightarrow (\text{conv3-64} \rightarrow \text{conv3-64}) \rightarrow (\text{conv3-64} \rightarrow \text{conv3-64})\Big] \rightarrow \Big[(\text{conv3-128} \rightarrow \text{conv3-128} \rightarrow \textcolor{red}{conv1-128}) \rightarrow (\text{conv3-128} \rightarrow \text{conv3-128}) \\
\rightarrow (\text{conv3-128} \rightarrow \text{conv3-128}) \rightarrow (\text{conv3-128} \rightarrow \text{conv3-128})\Big] \rightarrow \Big[(\text{conv3-256} \rightarrow \text{conv3-256} \rightarrow \textcolor{red}{conv1-256}) \rightarrow (\text{conv3-256} \rightarrow \text{conv3-256}) \rightarrow (\text{conv3-256} \rightarrow \text{conv3-256}) \rightarrow (\text{conv3-256} \rightarrow \text{conv3-256}) \\
\rightarrow (\text{conv3-256} \rightarrow \text{conv3-256}) \rightarrow (\text{conv3-256} \rightarrow \text{conv3-256})\Big] \rightarrow \Big[(\text{conv3-512} \rightarrow \text{conv3-512} \rightarrow \textcolor{red}{conv1-512}) \rightarrow (\text{conv3-512} \rightarrow \text{conv3-512}) \rightarrow (\text{conv3-512} \rightarrow \text{conv3-512})\Big] \rightarrow$ Output layer

\vspace{0.5em}\noindent
\emph{RESNET18}: Input layer $
\rightarrow \text{conv3-64} \rightarrow \Big[(\text{conv3-64} \rightarrow \text{conv3-64}) \rightarrow (\text{conv3-64} \rightarrow \text{conv3-64})\Big]
\rightarrow \Big[(\text{conv3-128} \rightarrow \text{conv3-128} \rightarrow \textcolor{red}{conv1-128}) \rightarrow (\text{conv3-128} \rightarrow \text{conv3-128})\Big]
\rightarrow \Big[(\text{conv3-256} \rightarrow \text{conv3-256} \rightarrow \textcolor{red}{conv1-256}) \rightarrow (\text{conv3-256} \rightarrow \text{conv3-256})\Big]
\rightarrow \Big[(\text{conv3-512} \rightarrow \text{conv3-512} \rightarrow \\
\textcolor{red}{conv1-512}) \rightarrow (\text{conv3-512} \rightarrow \text{conv3-512})\Big] \rightarrow$ Output layer

\begin{table}[t!]
    \centering
    \caption{Performance comparison between different mappings when fusing 2 RESNET models on the CIFAR10 dataset. Experiments were run with 5 different seeds. The term ``CLA woc" stands for CLA without the last layer constraint. The best performing mapping in columns $M_{\Ff}$ is in bold.}
    \scalebox{1.0}{
    \begin{tabular}{ccccc}
        \toprule
        Mapping & $M_A$ & $M_B$ & \multicolumn{2}{c}{$M_{\Ff}$} \\
        \cmidrule(lr){4-5}
        & & & Before finetuning & After finetuning \\
        \midrule
        Naive & \multirow{4}{*}{93.31 $\pm$ 0.13} & \multirow{4}{*}{92.92 $\pm$ 0.20} & 62.43 $\pm$ 3.08 & 93.59 $\pm$ 0.18 \\
        Random & & & 64.04 $\pm$ 3.01 & 93.61 $\pm$ 0.12 \\
        CLA woc & & & 65.58 $\pm$ 2.61 & 93.60 $\pm$ 0.14 \\
        CLA & & & \textbf{65.72 $\pm$ 2.33} & \textbf{93.67 $\pm$ 0.07} \\
        \bottomrule
    \end{tabular}
    }
    \label{table:ablation_studies_resnet}
    \vskip 0.2in
\end{table}

\vspace{0.5em}\noindent
\textbf{The optimal CLA.} Each RESNET model is also divided into 5 stages which are  enclosed in square brackets in the above representation. And each building block is enclosed in parentheses. As discussed in Section~\ref{subsec:cross_layer_alignment}, in each stage we find the optimal mapping for blocks instead of layers. Because each stage of RESNET18 has only two blocks, the first block is assigned to the first block while the second one is assigned to the last block in the same stage of RESNET34.

\vspace{0.5em}\noindent
\textbf{The effectiveness of the optimal mapping.} We compare the optimal CLA with three baseline mappings. The first baseline is the naive mapping of blocks with the same index, i.e., the $i^{th}$ block of RESNET18 is aligned to the $i^{th}$ block in the same stage of RESNET34. The second one is  random mapping which satisfies the first layer constraint. Because the total number of strictly increasing mappings is quite large, we only sample 5 random mappings for each random seed and take their average performance. The last mapping is the one that is obtained by solving the CLA problem but removing the last layer constraint. Table~\ref{table:ablation_studies_resnet} compares the classification accuracy of the fused model on different types of mapping. Before finetuning, using the optimal CLA outperforms the naive and random mappings with a wide margin of 3.29 and 1.68, respectively. Removing the last layer constraint decreases the performance slightly but still yields better performance than the other two. After finetuning, the optimal CLA produces slightly higher accuracy than the other mappings. In terms of baselines, three mappings have a comparative performance of approximately $93.60\%$.

\section{Additional experimental results}
\label{sec:addexp}
We first apply CLAFusion to the skill transfer experiment~\cite[~Section5.1]{singh2020model} but for heterogeneous neural networks. In the subsequent sections, we report the detailed results for different runs of experiments in the main text. Finally, we showcase the efficiency of CLAFusion in terms of computational resources.

\begin{figure}[t!]
    \begin{center}
        \begin{tabular}{cc}
            \widgraph{0.45\textwidth}{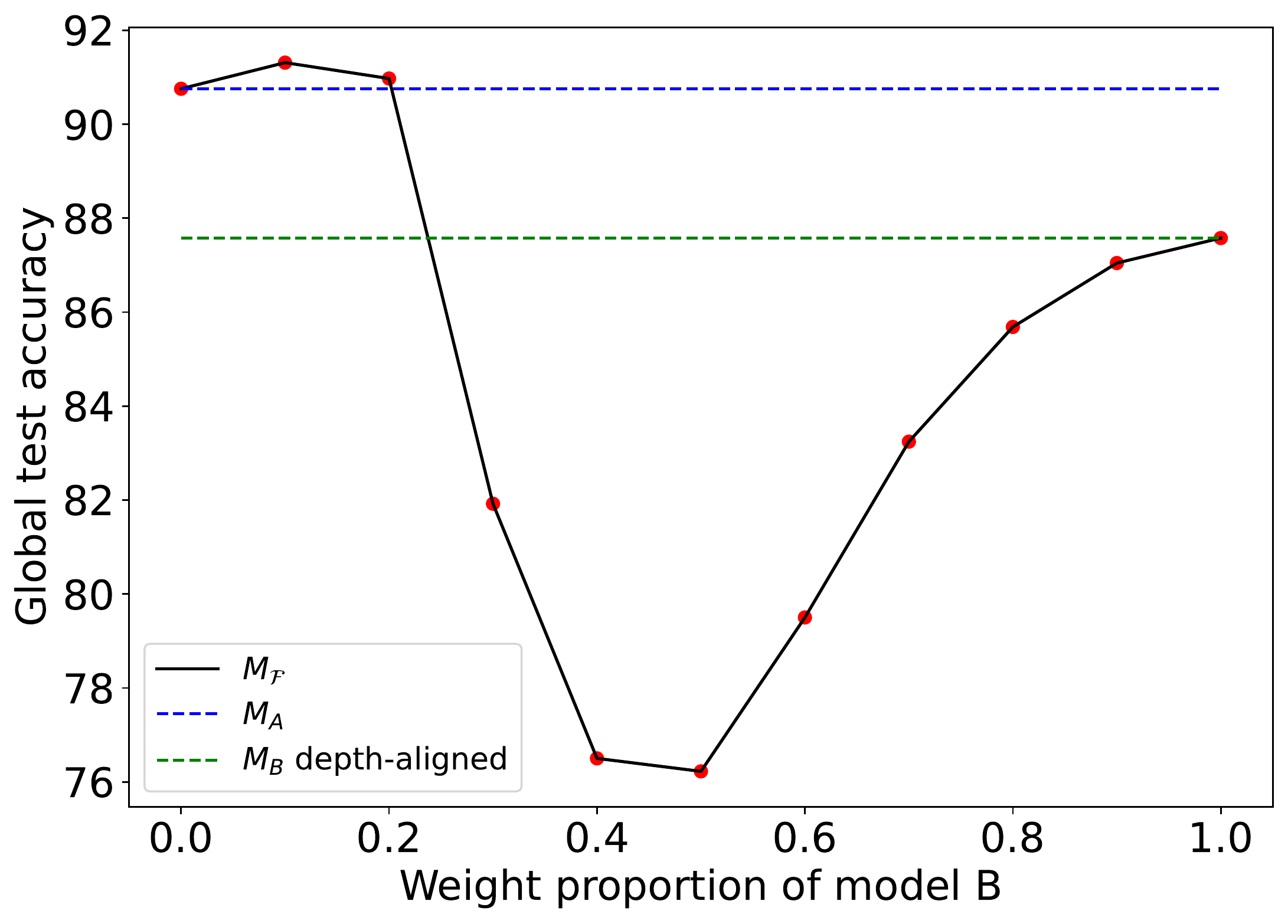} & \widgraph{0.46\textwidth}{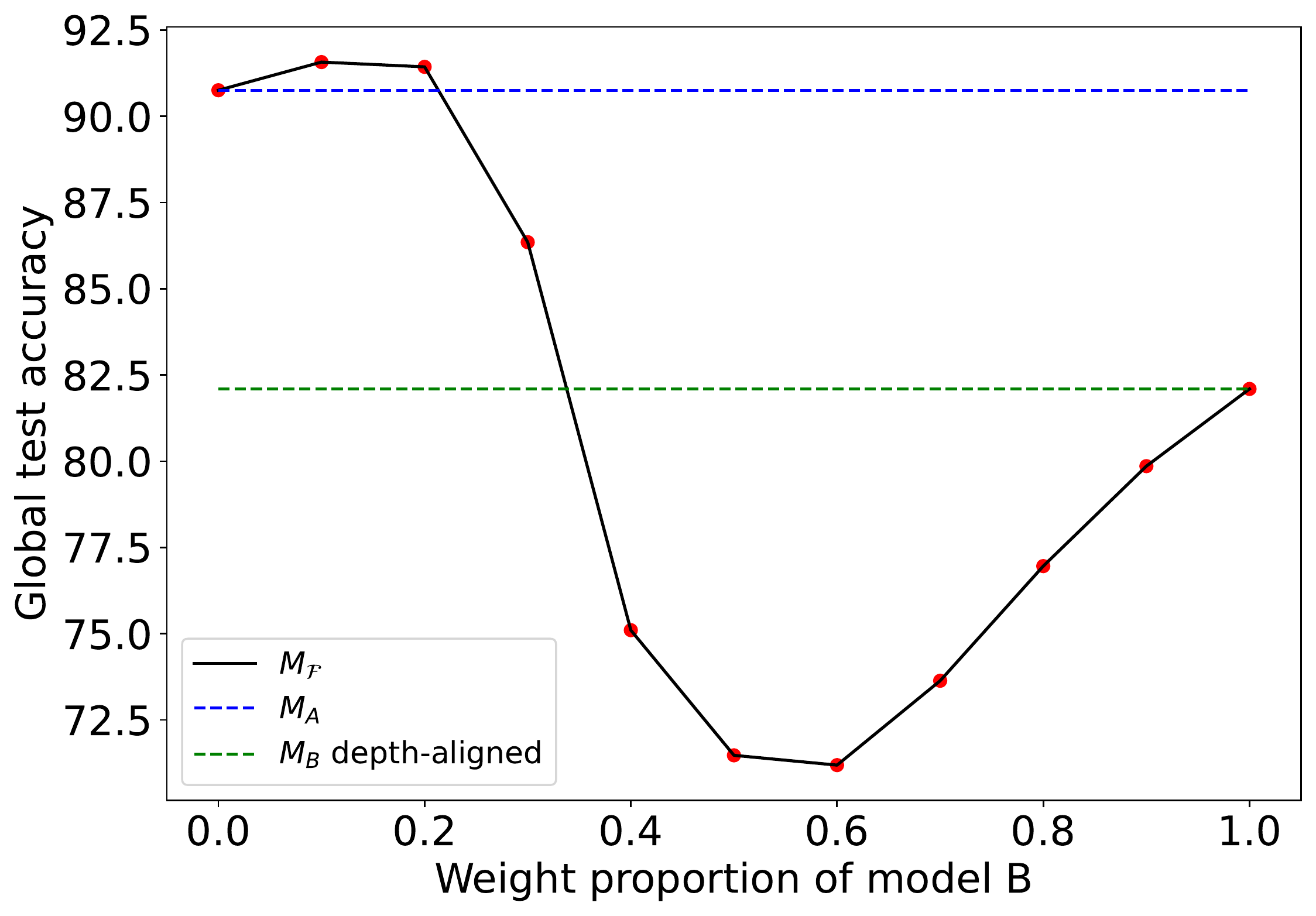} \\
            (a) Adding layers & (b) Merging layers (sum)
        \end{tabular}
    \end{center}
    \vskip -0.1in
    \caption{
        \footnotesize{The average performance of skill transfer using (a) adding layers and (b) merging layers as the layer balancing method. Experiments were run with 5 different seeds.}} 
    \label{fig:skill_transfer_comparison}
    \vskip 0.1in
\end{figure} 

\subsection{Skill transfer}
\label{subsec:addexp_skill_transfer}
\textbf{Settings.} In skill transfer, we aim to obtain a single model that can inherit both overall and specialized skills from the two individual models. We adopt the same heterogeneous data-split technique as in~\cite[Section~5.1]{singh2020model} for two MLPs A and B. The MNIST dataset is separated into two datasets. The dataset for model A has all images of label $4$ and $10\%$ images of other labels while that for model B contains the rest. We train an MLP with $3$ hidden layers of size $400, 200, 100$ for model A and an MLP with $4$ hidden layers of size $400, 200, 100, 50$ for model B. The pre-activation matrix is obtained by running inference for $400$ samples. We use the number of neurons as layer representation and adding layers method for balancing. The weight proportion of model A, which is referred to as $w_A$ (also known as fusion rate in Smith et al.~\cite{smith2017investigation}), is swept over 11 different values: $w_A = \{0.0, 0.1, \ldots, 1.0\}$.

\begin{table}[t!]
    \centering
    \caption{Performance comparison between two layer balancing methods on skill transfer task. Experiments were run with 5 different seeds. The best performing result in each row is in bold.}
    \scalebox{0.85}{
    \begin{tabu}{ccccccccccccc}
        \toprule
        & Seed & \multicolumn{11}{c}{$w_A$} \\
        \cmidrule(lr){3-13}
        & & 0.0 & 0.1 & 0.2 & 0.3 & 0.4 & 0.5 & 0.6 & 0.7 & 0.8 & 0.9 & 1.0 \\
        \midrule
        \multirow{6}{*}{Add} & 0 & 86.10 & 87.06 & \textbf{88.23} & 78.64 & 73.48 & 69.70 & 70.45 & 75.91 & 81.38 & 85.36 & 87.68 \\
        & 1 & 91.60 & \textbf{91.81} & 91.23 & 85.63 & 77.20 & 77.99 & 81.59 & 84.85 & 86.88 & 87.48 & 87.29 \\
        & 2 & 90.80 & 92.07 & \textbf{92.46} & 84.06 & 82.53 & 83.21 & 85.01 & 86.83 & 87.76 & 88.01 & 87.83 \\
        & 3 & 91.94 & \textbf{92.31} & 91.39 & 79.58 & 71.85 & 71.77 & 78.31 & 83.63 & 85.79 & 87.01 & 87.51 \\
        & 4 & \textbf{93.34} & 93.31 & 91.56 & 81.70 & 77.45 & 78.46 & 82.15 & 85.00 & 86.62 & 87.37 & 87.58 \\
        \cmidrule(lr){2-13}
        & Avg & 90.76 & \textbf{91.31} & 90.97 & 81.92 & 76.50 & 76.23 & 79.50 & 83.24 & 85.69 & 87.05 & 87.58 \\
        \midrule
        \multirow{6}{*}{Merge (sum)} & 0 & 86.10 & \textbf{87.19} & 87.10 & 78.51 & 63.66 & 56.74 & 53.31 & 53.88 & 57.65 & 66.55 & 76.35 \\
        & 1 & 91.60 & \textbf{92.23} & 91.83 & 89.08 & 78.69 & 76.04 & 77.34 & 81.44 & 84.87 & 86.39 & 86.60 \\
        & 2 & 90.80 & 91.98 & \textbf{92.30} & 86.00 & 81.27 & 79.42 & 80.09 & 82.95 & 85.46 & 85.61 & 85.04 \\
        & 3 & 91.94 & 92.66 & \textbf{92.76} & 88.13 & 71.65 & 65.90 & 64.20 & 67.39 & 72.91 & 76.06 & 77.40 \\
        & 4 & 93.34 & \textbf{93.80} & 93.18 & 90.03 & 80.24 & 79.27 & 81.01 & 82.52 & 83.91 & 84.68 & 85.09 \\
        \cmidrule(lr){2-13}
        & Avg & 90.76 & \textbf{91.57} & 91.43 & 86.35 & 75.10 & 71.47 & 71.19 & 73.64 & 76.96 & 79.86 & 82.10 \\
        \midrule
        \multirow{6}{*}{Merge (average)} & 0 & 86.10 & 87.55 & \textbf{88.65} & 85.23 & 69.70 & 60.47 & 57.45 & 59.01 & 63.90 & 75.26 & 84.55 \\
        & 1 & 91.60 & \textbf{92.16} & 91.62 & 89.28 & 79.00 & 75.20 & 76.80 & 81.53 & 85.00 & 86.75 & 87.10 \\
        & 2 & 90.80 & 91.96 & \textbf{92.35} & 88.33 & 81.11 & 79.69 & 80.60 & 83.80 & 86.28 & 86.64 & 86.42 \\
        & 3 & 91.94 & \textbf{92.74} & 92.60 & 87.63 & 74.14 & 68.91 & 68.37 & 74.89 & 80.75 & 84.04 & 85.61 \\
        & 4 & 93.34 & \textbf{93.83} & 92.80 & 89.48 & 80.96 & 78.68 & 80.50 & 83.19 & 84.82 & 86.11 & 86.56 \\
        \cmidrule(lr){2-13}
        & Avg & 90.76 & \textbf{91.65} & 91.60 & 87.99 & 76.98 & 72.59 & 72.74 & 76.48 & 80.15 & 83.76 & 86.05 \\
        \bottomrule
    \end{tabu}
    }
    \label{table:skill_transfer_details}
\end{table}

\vspace{0.5em}\noindent
\textbf{Results.} Figure~\ref{fig:skill_transfer_comparison}(a) shows that when the weight proportion of model B is small ($0.1$ or $0.2$), the fused model improves over both individual models. The reason is that the specialist model A has higher accuracy than the generalist model B. When distributing the weight proportion fairly ($0.5$), the fused model performs worst due to the effect of heterogeneous training data. The fused model has the best accuracy of $91.31\%$, which is higher than those of models A ($90.76\%$) and B ( $87.58\%$). This claims that the knowledge has been transferred successfully to the fused model without retraining. Furthermore, the detailed results of skill transfer for 5 different random seeds are reported in Table~\ref{table:skill_transfer_details}. The specialist model A generally has higher accuracy than the generalist model B seeds because it has been trained on all 10 labels. In 4 out of 5 seeds, the fused model improves over both individual models when the weight proportion of model B is small ($w_A = \{0.1, 0.2\}$). 

\vspace{0.5em}\noindent
\textbf{Comparison between two layer balancing methods.} For comparison purposes, we further conduct the same procedure but replace adding layers by merging layers (the sign of sum and the average of sign estimation). Similar to the previous experiment, the fused model attains the best performance when the weight proportion of model B is small ($w_A = \{0.1, 0.2\}$). Figure~\ref{fig:skill_transfer_comparison} illustrates the average performance of two layer balancing methods on the skill transfer task. When averaging over 5 random seeds, both methods perform best if $w_A = 0.1$. The best accuracy of merging layers $(91.57, 91.65)$ is slightly higher than that of adding layers $(91.31)$ even though the accuracy of $M_B$ depth-aligned drops from $87.58$ to $82.10$ or $86.05$.

\begin{table}[t!]
    \centering
    \caption{The results of finetuning RESNET from different choices of initialization across 5 different seeds. The best performing initialization in each row is in bold.}
    \scalebox{0.75}{
    \begin{tabu}{ccccccccccc}
        \toprule
        Dataset & Seed & $M_A$ & $M_B$ & Ensemble & $M_{\Ff}$ &  \multicolumn{5}{c}{Finetune} \\
        \cmidrule(lr){7-11}
        & & & & Learning & & $M_A$ & $M_B$ & $M_{\Ff}$ & $M_B$ depth-aligned & Random RESNET34  \\
        \midrule
        CIFAR10 & 40 & 93.42 & 93.21 & 94.10 & 67.83 & 93.50 & 93.47 & \textbf{93.61} & 93.47 & 91.95 \\
        & 41 & 93.37 & 92.90 & 93.81 & 63.66 & 93.61 & 93.26 & \textbf{93.74} & 93.32 & 91.93 \\
        & 42 & 93.31 & 93.05 & 93.94 & 68.30 & 93.61 & 93.36 & \textbf{93.71} & 93.43 & 92.25 \\
        & 43 & 93.39 & 92.78 & 93.71 & 66.45 & 93.50 & 93.30 & \textbf{93.73} & 93.00 & 91.76 \\
        & 44 & 93.06 & 92.65 & 93.49 & 62.35 & 93.37 & 93.05 & \textbf{93.55} & 92.97 & 92.12 \\
        \cmidrule(lr){2-11}
        & Avg & 93.31 & 92.92 & 93.81 & 65.72 & 93.52 & 93.29 & \textbf{93.67} & 93.24 & 92.00 \\
        \midrule
        CIFAR100 & 40 & 66.08 & 65.19 & 68.44 & 29.44 & 66.73 & 66.47 & \textbf{67.71} & 66.23 & 64.73 \\
        & 41 & 66.21 & 65.22 & 68.58 & 28.92 & 67.32 & 66.33 & \textbf{67.54} & 66.45 & 64.24 \\
        & 42 & 65.89 & 65.04 & 68.39 & 26.62 & 66.77 & 65.91 & \textbf{67.16} & 66.35 & 64.66 \\
        & 43 & 65.85 & 66.17 & 68.98 & 27.35 & 67.12 & 66.35 & \textbf{68.05} & 67.23 & 65.67 \\
        & 44 & 65.61 & 65.03 & 68.05 & 27.31 & 66.51 & 65.90 & \textbf{67.50} & 66.16 & 63.65 \\
        \cmidrule(lr){2-11}
        & Avg & 65.93 & 65.33 & 68.49 & 27.93 & 66.87 & 66.19 & \textbf{67.59} & 66.48 & 64.59 \\
        \midrule
        Tiny-ImageNet & 40 & 54.29 & 53.13 & 58.27 & 17.22 & 54.50 & 53.71 & \textbf{54.75} & 53.77 & 53.91 \\
        & 41 & 54.81 & 53.74 & 59.44 & 17.84 & 55.20 & 53.17 & \textbf{55.23} & 53.24 & 53.26 \\
        & 42 & 55.37 & 53.74 & 59.44 & 17.84 & 55.87 & 54.26 & \textbf{56.00} & 54.60 & 54.44 \\
        & 43 & 54.83 & 53.62 & 59.13 & 14.82 & 55.27 & 53.91 & \textbf{55.96} & 54.10 & 54.87 \\
        & 44 & 54.91 & 53.28 & 58.31 & 18.33 & 54.99 & 54.02 & \textbf{55.35} & 53.98 & 54.32 \\
        \cmidrule(lr){2-11}
        & Avg & 54.84 & 53.29 & 58.70 & 17.19 & 55.17 & 53.81 & \textbf{55.46} & 53.94 & 54.16 \\
        \bottomrule
    \end{tabu}
    }
    \label{table:fuse_two_models_details}
\end{table}

\begin{table}[t!]
    \centering
    \caption{The results of fusing and finetuning 4 RESNET from different choices of initialization across 5 different seeds. The best performing initialization in each row is in bold.}
    \scalebox{0.85}{
    \begin{tabu}{ccccccc}
        \toprule
        Dataset & Seeds & Pre-trained models & Ensemble & $M_{\Ff}$ &  \multicolumn{2}{c}{Finetune} \\
        \cmidrule(lr){6-7}
        & & (RESNET34, RESNET18) x2 & Learning & & Pre-trained models & $M_{\Ff}$ \\
        \midrule
        CIFAR10 & 40, 41 & 93.42, 93.21, 93.37, 92.90 & 94.36 & 34.55 & 93.50, 93.47, 93.61, 93.26 & \textbf{93.97} \\
        & 41, 42 & 93.37, 92.90, 93.31, 93.05 & 94.20 & 32.35 & 93.61, 93.26, 93.61, 93.36 & \textbf{93.72} \\
        & 42, 43 & 93.31, 93.05, 93.39, 92.78 & 94.10 & 34.01 & 93.61, 93.36, 93.50, 93.30 & \textbf{93.81} \\
        & 43, 44 & 93.39, 92.78, 93.06, 92.65 & 94.11 & 34.93 & 93.50, 93.30, 93.37, 93.05 & \textbf{93.73} \\
        & 44, 45 & 93.06, 92.65, 92.66, 92.76 & 94.06 & 22.98 & 93.37, 93.05, 92.94, 93.04 & \textbf{93.56} \\
        \cmidrule(lr){2-7}
        & Avg & 93.31, 92.92, 93.16, 92.83 & 94.17 & 31.76 & 93.52, 93.29, 93.41, 93.20 & \textbf{93.76} \\
        \midrule
        CIFAR100 & 40, 41 & 66.08, 65.19, 66.21, 65.22 & 70.24 & 8.28 & 66.73, 66.47, 67.23, 66.33 & \textbf{68.89} \\
        & 41, 42 & 66.21, 65.22, 65.89, 65.04 & 70.50 & 8.02 & 67.23, 66.33, 66.77, 65.91 & \textbf{68.70} \\
        & 42, 43 & 65.89, 65.04, 65.85, 66.17 & 70.46 & 7.75 & 66.77, 65.91, 67.12, 66.35 & \textbf{68.77} \\
        & 43, 44 & 65.85, 66.17, 65.61, 65.03 & 70.16 & 8.51 & 67.12, 66.35, 66.51, 65.90 & \textbf{69.08} \\
        & 44, 45 & 65.61, 65.03, 66.27, 64.92 & 69.98 & 9.77 & 66.51, 65.90, 66.24, 65.29 & \textbf{69.14} \\
        \cmidrule(lr){2-7}
        & Avg & 65.93, 65.33, 65.97, 65.28 & 70.27 & 8.47 & 66.87, 66.19, 66.77, 65.96 & \textbf{68.92} \\
        \midrule
        Tiny-ImageNet & 40, 41 & 54.29, 53.13, 54.81, 53.74 & 61.28 & 3.24 & 54.50, 53.71, 55.20, 53.17 & \textbf{55.67} \\
        & 41, 42 & 54.81, 53.74, 59.44, 55.37 & 61.47 & 4.70 & 55.20, 53.17, 55.87, 54.26 & \textbf{56.35} \\
        & 42, 43 & 55.37, 53.74, 54.83, 53.62 & 61.50 & 3.11 & 55.87, 54.26, 55.27, 53.91 & \textbf{56.14} \\
        & 43, 44 & 54.83, 53.62, 54.91, 53.28 & 61.23 & 4.13 & 55.27, 53.91, 54.99, 54.02 & \textbf{56.32} \\
        & 44, 45 & 54.91, 53.28, 54.65, 52.81 & 60.92 & 5.06 & 54.99, 54.02, 55.11, 53.44 & \textbf{55.86} \\
        \cmidrule(lr){2-7}
        & Avg & 54.84, 53.29, 54.91, 53.23 & 61.28 & 4.05 & 55.17, 53.81, 55.29, 53.76 & \textbf{56.07} \\
        \bottomrule
    \end{tabu}
    }
    \label{table:fuse_four_models_details}
\end{table}

\subsection{Generate an efficient initialization}
\label{subsec:addexp_eff_init}
\textbf{Results.} Finetuning from the fused model always improves the accuracy over finetuning from pre-trained models as illustrated in Table~\ref{table:fuse_two_models_details}. In comparison with other initializations, it yields the best performance in all 5 seeds on both datasets. $M_B$ depth-aligned, which is the result of NET2NET~\cite{chen2015net2net} operation, performs better than finetuning from scratch but only leads to comparable performance as finetuning from $M_B$. 

\vspace{0.5em}\noindent
\textbf{Pre-trained models for multiple model cases.} For 4 and 6 models, we train two and three pairs of RESNET34 and RESNET18 models, respectively. To utilize pre-trained models in the two model scenario, we simply group multiple pre-trained model pairs at different seeds. Specifically, in Table~\ref{table:fuse_four_models_details}, the first pair (the first and second pre-trained models) was trained at seed 40 and the second pair (the third and fourth pre-trained models) was trained at seed 41. In all pairs, the RESNET34 model is always put before the RESNET18 model. Their accuracies can be found in Table~\ref{table:fuse_two_models_details} except for seeds 45 and 46, which are newly trained for this setting.

\vspace{0.5em}\noindent
\textbf{CLAFusion for more than 2 models.} Tables~\ref{table:fuse_four_models_details} and~\ref{table:fuse_six_models_details} detail the results in the multiple-model scenario. Compared to the two-model scenario, fusing 4 RESNET results in the initialization with a significant decrease in the accuracy due to the increasing difficulty of the task. However, there is an increase in the final performance after finetuning, with a figure growing from $93.67$ to $93.76$ on the CIFAR10 dataset, from $67.59$ to $68.92$ on the CIFAR100 dataset, and from $55.46$ to $56.07$ on the Tiny-ImageNet dataset. In addition, CLAFusion produces the best initialization in all 5 runs. For the 6-model case, CLAFusion also yields more favorable initialization in every run on CIFAR10, CIFAR100, and Tiny-ImageNet datasets. It is worth mentioning that increasing the number of pre-trained models increase the performance of ensemble learning and the fused model after finetuning. This outcome is expected since both methods combine knowledge from pre-trained models that were trained at different random seeds.

\begin{table}[t!]
    \centering
    \caption{The results of fusing and finetuning 6 RESNET from different choices of initialization across 5 different seeds. The best performing initialization in each row is in bold.}
    \scalebox{0.68}{
    \begin{tabu}{ccccccc}
        \toprule
        Dataset & Seeds & Pre-trained models & Ensemble & $M_{\Ff}$ &  \multicolumn{2}{c}{Finetune} \\
        \cmidrule(lr){6-7}
        & & (RESNET34, RESNET18) x3 & Learning & & Pre-trained models & $M_{\Ff}$ \\
        \midrule
        CIFAR10 & 40, 41, 42 & 93.42, 93.21, 93.37, 92.90, 93.31, 93.05 & 94.28 & 26.32 & 93.50, 93.47, 93.61, 93.26, 93.61, 93.36 & \textbf{93.84} \\
        & 41, 42, 43 & 93.37, 92.90, 93.31, 93.05, 93.39, 92.78 & 94.19 & 24.22 & 93.61, 93.26, 93.61, 93.36, 93.50, 93.30 & \textbf{93.84} \\
        & 42, 43, 44 & 93.31, 93.05, 93.39, 92.78, 93.06, 92.65 & 94.13 & 24.37 & 93.61, 93.36, 93.50, 93.30, 93.37, 93.05 & \textbf{93.63} \\
        & 43, 44, 45 & 93.39, 92.78, 93.06, 92.65, 92.66, 92.76 & 94.14 & 23.78 & 93.50, 93.30, 93.37, 93.05, 92.94, 93.04 & \textbf{93.69} \\
        & 44, 45, 46 & 93.06, 92.65, 92.66, 92.76, 93.46, 93.05 & 94.36 & 20.08 & 93.37, 93.05, 92.94, 93.04, 93.41, 93.34 & \textbf{93.89} \\
        \cmidrule(lr){2-7}
        & Avg & 93.31, 92.92, 93.16, 92.83, 93.18, 92.86 & 94.22 & 23.75 & 93.52, 93.29, 93.41, 93.20, 93.49, 93.24 & \textbf{93.78} \\
        \midrule
        CIFAR100 & 40, 41, 42 & 66.08, 65.19, 66.21, 65.22, 65.89, 65.04 & 70.47 & 4.14 & 66.73, 66.47, 67.23, 66.33, 66.77, 65.91 & \textbf{68.80} \\
        & 41, 42, 43 & 66.21, 65.22, 65.89, 65.04, 65.85, 66.17 & 70.83 & 4.63 & 67.23, 66.33, 66.77, 65.91, 67.12, 66.35 & \textbf{69.34} \\
        & 42, 43, 44 & 65.89, 65.04, 65.85, 66.17, 65.61, 65.03 & 70.66 & 4.22 & 66.77, 65.91, 67.12, 66.35, 66.51, 65.90 & \textbf{69.72} \\
        & 43, 44, 45 & 65.85, 66.17, 65.61, 65.03, 66.27, 64.92 & 70.53 & 4.87 & 67.12, 66.35, 66.51, 65.90, 66.24, 65.29 & \textbf{69.56} \\
        & 44, 45, 46 & 65.61, 65.03, 66.27, 64.92, 65.82, 65.34 & 70.63 & 4.09 & 66.51, 65.90, 66.24, 65.29, 66.18, 66.16 & \textbf{69.57} \\
        \cmidrule(lr){2-7}
        & Avg & 65.93, 65.33, 65.97, 65.28, 65.89, 65.30 & 70.62 & 4.39 & 66.87, 66.19, 66.77, 65.96, 66.56, 65.92 & \textbf{69.40} \\
        \midrule
        Tiny-ImageNet & 40, 41, 42 & 54.29, 53.13, 54.81, 53.74, 59.44, 55.37 & 62.51 & 2.12 & 54.50, 53.71, 55.20, 53.17, 55.87, 54.26 & \textbf{56.15} \\
        & 41, 42, 43 & 54.81, 53.74, 59.44, 55.37, 54.83, 53.62 & 62.48 & 3.22 & 55.20, 53.17, 55.87, 54.26, 55.27, 53.91 & \textbf{57.05} \\
        & 42, 43, 44  & 55.37, 53.74, 54.83, 53.62, 54.91, 53.28 & 62.47 & 1.77 & 55.87, 54.26, 55.27, 53.91, 54.99, 54.02 & \textbf{56.51} \\
        & 43, 44, 45 & 54.83, 53.62, 54.91, 53.28, 54.65, 52.81 & 62.29 & 2.36 & 55.27, 53.91, 54.99, 54.02, 55.11, 53.44 & \textbf{56.44} \\
        & 44, 45, 46 & 54.91, 53.28, 54.65, 52.81, 53.39, 53.92 & 61.87 & 2.87 & 54.99, 54.02, 55.11, 53.44, 54.05, 54.55 & \textbf{56.21} \\
        \cmidrule(lr){2-7}
        & Avg & 54.84, 53.29, 54.91, 53.23, 54.63, 53.47 & 62.32 & 2.47 & 55.17, 53.81, 55.29, 53.76, 55.06, 54.04 & \textbf{56.47} \\
        \bottomrule
    \end{tabu}
    }
    \label{table:fuse_six_models_details}
\end{table}

\begin{table}[t!]
    \centering
    \caption{The results of model transfer from $M_B$ (RESNET18) to $M_A$ (RESNET34) across 5 different seeds. The best performing method in each row is in bold.}
    \scalebox{0.75}{
    \begin{tabular}{ccccccc}
        \toprule
        Dataset & Method & Seed & HMT & HMT + Naive avg & HMT + OTFusion & HMT + CLA + OTFusion \\
        \midrule
        CIFAR10 & Transfer & 40 & 9.55 & 37.80 & 20.17 & \textbf{62.94} \\
        & & 41 & 10.13 & 37.66 & 17.66 & \textbf{62.27} \\
        & & 42 & 10.45 & 43.44 & 19.25 & \textbf{60.86} \\
        & & 43 & 12.44 & 39.24 & 15.20 & \textbf{65.74} \\
        & & 44 & 11.37 & 29.33 & 19.30 & \textbf{54.63} \\
        \cmidrule(lr){4-7}
        & & Avg & 10.79 & 37.49 & 18.32 & \textbf{61.29} \\
        \cmidrule(lr){2-7}
        & Transfer + Finetune & 40 & 93.29 & 93.63 & 93.61 & \textbf{93.75} \\
        & & 41 & 93.14 & 93.28 & 93.69 & \textbf{94.09} \\
        & & 42 & 93.08 & 93.16 & \textbf{93.78} & 93.68 \\
        & & 43 & 93.20 & 93.27 & 93.80 & \textbf{93.92} \\
        & & 44 & 92.77 & 93.58 & 93.41 & \textbf{93.84} \\
        \cmidrule(lr){4-7}
        & & Avg & 93.10 & 93.38 & 93.66 & \textbf{93.86} \\
        \midrule
        CIFAR100 & Transfer & 40 & 2.98 & 9.69 & 7.32 & \textbf{12.69} \\
        & & 41 & 2.33 & 10.92 & 4.38 & \textbf{14.49} \\
        & & 42 & 3.03 & \textbf{12.90} & 4.13 & 7.33 \\
        & & 43 & 3.17 & 11.44 & 5.65 & \textbf{15.35} \\
        & & 44 & 2.00 & 9.00 & 4.22 & \textbf{12.92} \\
        \cmidrule(lr){4-7}
        & & Avg & 2.70 & 10.79 & 5.14 & \textbf{12.56} \\
        \cmidrule(lr){2-7}
        & Transfer + Finetune & 40 & 64.78 & 65.96 & 67.10 & \textbf{67.74} \\
        & & 41 & 65.25 & 66.39 & 67.37 & \textbf{67.95} \\
        & & 42 & 65.15 & 66.56 & 67.22 & \textbf{67.51} \\
        & & 43 & 65.53 & 66.71 & 67.44 & \textbf{68.13} \\
        & & 44 & 65.06 & 66.46 & 67.00 & \textbf{67.80} \\
        \cmidrule(lr){4-7}
        & & Avg & 65.15 & 66.42 & 67.23 & \textbf{67.83} \\
        \bottomrule
    \end{tabular}
    }
    \label{table:model_transfer_details}
\end{table}

\subsection{Heterogeneous model transfer}
\label{subsec:addexp_hmt}
\textbf{Settings.} The finetuning hyperparameters are similar to those for finetuning RESNET models in Appendix~\ref{subsec:addexp_eff_init}. However, we observe that retraining at the initial learning rate of 0.1 in some cases causes the target model after transferring to diverge. Hence, we additionally retrain at the initial learning rate of 0.05 and report the best performing one between two initial learning rates.

\vspace{0.5em}\noindent
\textbf{Results.} Table~\ref{table:model_transfer_details} details the results of 4 different model transfer methods. On the CIFAR10 dataset, combining CLA and OTFusion with HMT always results in a great boost, at least $16.27$ (seed 42), in the accuracy of the target model after transferring. After the finetuning phase, it gives the best performance in 4 out of 5 random seeds. In terms of the CIFAR100 dataset, incorporating CLAFusion into HMT leads to the best performance in 5 out of 5 runs. 

\subsection{Teacher-student fusion}
\label{subsec:addexp_teacher_student_fusion}
\textbf{Settings.} We conduct an additional setting, which is fusing teacher and student RESNET models on the CIFAR100 dataset. Teacher model A has RESNET34 architecture while model B has RESNET18 architecture with a half number of channels at each layer. The architectures of $M_A, M_B,$ and $M_{\Ff}$, which are in turn RESNET34, RESNET18 half, and RESNET34 half, can be represented as follows.

\vspace{0.5em}\noindent
\emph{RESNET34}: Input layer $
\rightarrow \text{conv3-64} \rightarrow \text{conv3-64} \rightarrow \text{conv3-64} \rightarrow \text{conv3-64} \rightarrow \text{conv3-64} \rightarrow \text{conv3-64} \rightarrow \text{conv3-64}
\rightarrow \text{conv3-128} \rightarrow \text{conv3-128} \rightarrow \textcolor{red}{conv1-128} \rightarrow \text{conv3-128} \rightarrow \text{conv3-128} \rightarrow \text{conv3-128} \rightarrow \text{conv3-128} \rightarrow \text{conv3-128} \rightarrow \text{conv3-128} 
\rightarrow \text{conv3-256} \rightarrow \text{conv3-256} \rightarrow \textcolor{red}{conv1-256} \rightarrow \text{conv3-256} \rightarrow \text{conv3-256} \rightarrow \text{conv3-256} \rightarrow \text{conv3-256} \rightarrow \text{conv3-256} \rightarrow \text{conv3-256} \rightarrow \text{conv3-256} \rightarrow \text{conv3-256} \rightarrow \text{conv3-256} \rightarrow \text{conv3-256} 
\rightarrow \text{conv3-512} \rightarrow \text{conv3-512} \rightarrow \textcolor{red}{conv1-512} \rightarrow \text{conv3-512} \rightarrow \text{conv3-512} \rightarrow \text{conv3-512} \rightarrow \text{conv3-512} \rightarrow$ Output layer

\vspace{0.5em}\noindent
\emph{RESNET18 half}: Input layer $
\rightarrow \text{conv3-32} \rightarrow \text{conv3-32} \rightarrow \text{conv3-32} \rightarrow \text{conv3-32} \rightarrow \text{conv3-32} 
\rightarrow \text{conv3-64} \rightarrow \text{conv3-64} \rightarrow \textcolor{red}{conv1-64} \rightarrow \text{conv3-64} \rightarrow \text{conv3-64} 
\rightarrow \text{conv3-128} \rightarrow \text{conv3-128} \rightarrow \textcolor{red}{conv1-128} \rightarrow \text{conv3-128} \rightarrow \text{conv3-128} 
\rightarrow \text{conv3-256} \rightarrow \text{conv3-256} \rightarrow \textcolor{red}{conv1-256} \rightarrow \text{conv3-256} \rightarrow \text{conv3-256} \rightarrow$ Output layer

\vspace{0.5em}\noindent
\emph{RESNET34 half}:  Input layer $
\rightarrow \text{conv3-32} \rightarrow \text{conv3-32} \rightarrow \text{conv3-32} \rightarrow \text{conv3-32} \rightarrow \text{conv3-32} \rightarrow \text{conv3-32} \rightarrow \text{conv3-32}
\rightarrow \text{conv3-64} \rightarrow \text{conv3-64} \rightarrow \textcolor{red}{conv1-64} \rightarrow \text{conv3-64} \rightarrow \text{conv3-64} \rightarrow \text{conv3-64} \\
\rightarrow \text{conv3-64} \rightarrow \text{conv3-64} \rightarrow \text{conv3-64} 
\rightarrow \text{conv3-128} \rightarrow \text{conv3-128} \rightarrow \textcolor{red}{conv1-128} \rightarrow \text{conv3-128} \rightarrow \text{conv3-128} \rightarrow \text{conv3-128} \rightarrow \text{conv3-128} \rightarrow \text{conv3-128} \rightarrow \text{conv3-128} \rightarrow \text{conv3-128} \rightarrow \text{conv3-128} \rightarrow \text{conv3-128} \rightarrow \text{conv3-128} 
\rightarrow \text{conv3-256} \rightarrow \text{conv3-256} \rightarrow \textcolor{red}{conv1-256} \rightarrow \text{conv3-256} \rightarrow \text{conv3-256} \rightarrow \text{conv3-256} \rightarrow \text{conv3-256} \rightarrow$ Output layer

\vspace{0.5em}\noindent
Other settings for running CLAFusion and finetuning are similar to experiments in Section~\ref{subsec:eff_init}.

\begin{table}[t!]
    \centering
    \caption{Finetuning teacher-student models across 5 different seeds. The best performing student model in each row is in bold. The results of VGG reported in Table~\ref{table:teacher_student_fusion} are run with the random seed of 42. }
    \scalebox{0.9}{
    \begin{tabular}{cccccccccc}
        \toprule
        Dataset + & Seed & $M_A$ & $M_B$ & Ensemble & $M_{\Ff}$ &  \multicolumn{4}{c}{Retrain} \\
        \cmidrule(lr){7-10}
        Architecture & & & & Learning & & $M_A$ & $M_B$ & $M_{\Ff}$ & $M_B$ depth-aligned\\
        \midrule
        CIFAR10 + & 40 & 92.71 & 89.68 & 92.53 & 81.66 & 92.67 & 89.81 & \textbf{90.64} & 90.54 \\
        VGG & 41 & 92.34 & 89.61 & 92.68 & 82.59 & 92.50 & 89.57 & \textbf{90.59} & 90.42 \\
        & 42 & 92.70 & 89.92 & 92.86 & 82.66 & 92.65 & 89.89 & \textbf{90.96} & 90.84 \\
        & 43 & 92.79 & 89.78 & 92.82 & 81.79 & 92.77 & 90.10 & 90.55 & \textbf{90.61} \\
        & 44 & 92.70 & 89.62 & 92.51 & 81.43 & 92.56 & 89.46 & \textbf{90.44} & 90.19 \\
        \cmidrule(lr){3-10}
        & Avg & 92.65 & 89.72 & 92.68 & 82.03 & 92.63 & 89.77 & \textbf{90.64} & 90.52 \\
        \midrule
        CIFAR100 + & 40 & 66.08 & 61.68 & 67.14 & 28.72 & 66.73 & 63.10 & \textbf{64.89} & 63.54 \\
        RESNET & 41 & 66.21 & 61.57 & 67.11 & 27.98 & 67.23 & 62.95 & \textbf{64.59} & 62.94 \\
        & 42 & 65.89 & 61.36 & 67.53 & 29.31 & 66.77 & 62.97 & \textbf{65.10} & 62.48 \\
        & 43 & 65.85 & 61.56 & 67.14 & 28.03 & 67.12 & 62.23 & \textbf{63.83} & 62.96 \\
        & 44 & 65.61 & 61.46 & 66.94 & 28.30 & 66.51 & 62.59 & \textbf{63.85} & 62.84 \\
        \cmidrule(lr){3-10}
        & Avg & 65.93 & 61.53 & 67.17 & 28.47 & 66.87 & 62.77 & \textbf{64.45} & 62.95 \\
        \bottomrule
    \end{tabular}
    }
    \label{table:teacher_student_details}
\end{table}

\vspace{0.5em}\noindent
\textbf{Results.} Table~\ref{table:teacher_student_details} illustrates the finetuning result of teacher and student models across $5$ different seeds. The fused model $M_{\Ff}$ is the most productive initialization of student models in at least 4 out of 5 seeds. Retraining the fused model ($M_{\Ff}$) always yields higher accuracy than continuing training the student model ($M_B$), leading to average improvements of 0.87 and 1.68 on CIFAR10 and CIFAR100 datasets, respectively.

\vspace{0.5em}\noindent
\textbf{Discussion.} At first glance, the first two steps of our approach may resemble the NET2NET operations. Although both approaches increase the depth of the student network and use the deeper student as a good initialization, there are two major differences between ours and NET2NET. Firstly, we present a systematic way of finding the location to add the identity mappings for different types of network architectures while NET2NET is manually designed for a specific type of network. Secondly, we do not enlarge the width of the student network (equivalently, use only NET2DEEPERNET operation) so that the student network remains more compact than the teacher network.

\begin{table}[t!]
    \centering
    \caption{Distillation results for different temperatures. For each type of architecture, the entity in the first five rows is the best accuracy obtained by varying the loss-weight factor while the last two rows report the best and average performance across 5 different temperatures. The best performing initialization in each row is in bold.}
    \scalebox{0.85}{
    \begin{tabular}{ccccccc}
        \toprule
        Dataset + & Temperature & \multicolumn{5}{c}{Distillation initialization} \\
        \cmidrule(lr){3-7}
        Architecture & T & Random $M_B$ & $M_B$ & $M_{\Ff}$ & $M_B$ depth-aligned & Random $M_{\Ff}$ \\
        \midrule
        CIFAR10 + & 20 & 88.99 (0.70) & 90.94 (0.70) & \textbf{91.29} (0.10) & 91.14 (0.70) & 88.20 (0.10) \\
        VGG & 10 & 89.10 (0.70) & 90.97 (0.70) & \textbf{91.23} (0.50) & \textbf{91.23} (0.10) & 88.84 (0.50) \\
        & 8 & 88.91 (0.70) & 90.98 (0.70) & \textbf{91.43} (0.70) & 91.24 (0.70) & 88.76 (0.50) \\
        & 4 & 87.89 (0.70) & 90.81 (0.99) & \textbf{91.14} (0.50) & 91.10 (0.50) & 87.65 (0.05) \\
        & 1 & 87.14 (0.10) & 89.97 (0.10) & 90.69 (0.05) & \textbf{90.73} (0.05) & 87.04 (0.05) \\
        \cmidrule(lr){2-7}
        & Best & 89.10 & 90.98 & \textbf{91.43} & 91.24 & 88.84 \\
        & Avg & 88.41 & 90.73 & \textbf{91.16} & 91.09 & 88.10 \\
        \midrule
        CIFAR100 + & 20 & 63.19 (0.50) & 65.87 (0.95) & \textbf{66.56} (0.70) & 66.04 (0.70) & 64.37 (0.50) \\
        RESNET & 10 & 63.99 (0.50) & 66.20 (0.95) & \textbf{66.79} (0.70) & 66.31 (0.99) & 64.84 (0.95) \\
        & 8 & 63.01 (0.50) & 66.07 (0.99) & \textbf{66.50} (0.70) & 66.08 (0.99) & 64.63 (0.70) \\
        & 4 & 62.41 (0.50) & 64.54 (0.50) & \textbf{65.26} (0.70) & 64.52 (0.50) & 62.79 (0.50) \\
        & 1 & 61.64 (0.10) & 63.60 (0.10) & \textbf{64.43} (0.05) & 63.94 (0.50) & 62.01 (0.05) \\
        \cmidrule(lr){2-7}
        & Best & 63.99 & 66.20 & \textbf{66.79} & 66.31 & 64.84 \\
        & Avg & 62.85 & 65.26 & \textbf{65.92} & 65.38 & 63.73 \\
        \bottomrule
    \end{tabular}
    }
    \label{table:knowledge_distillation_details}
\end{table}

\subsection{Knowledge distillation}
\label{subsec:knowledge_distillation}
In this experiment, we apply knowledge distillation to pre-trained models of seed 42 in Table~\ref{table:teacher_student_details}. For a fair comparison, the hyperparameters for training student models are identical to finetune hyperparameters in Section~\ref{subsec:teacher_student_fusion}.

\vspace{0.5em}\noindent
\textbf{Results.} The results of distilling the teacher model into the student model are given in Table~\ref{table:knowledge_distillation_details}. Knowledge distillation from the fused model leads to the best accuracy in no fewer than 4 out of 5 temperatures. In addition, both its average and best performance are the highest among all initializations. This suggests that CLAFusion can also be used to generate an efficient initialization for pre-trained distillation. Moving onto other initialization, $M_B$ depth-aligned, which is obtained from the NET2NET operation, is the second-best choice, followed by the pre-trained student model $M_B$. Distilling knowledge into either randomly initialized student models, however, yields a much lower accuracy. This result is consistent with the findings in Turc et al.~\cite{turc2019well}.

\begin{table*}[t!]
    \centering
    \caption{The number of parameters and inference time on the CIFAR10 dataset for different experiments. The inference time is the time taken in seconds to run inference on the whole test dataset.}
    \scalebox{0.65}{
    \begin{tabular}{llclcclcc}
        \toprule
        Experiment & \multicolumn{2}{c}{Architecture} & \multicolumn{3}{c}{\# Params} &\multicolumn{3}{c}{Inference time} \\
        \cmidrule(lr){2-3} \cmidrule(lr){4-6} \cmidrule(lr){7-9}
        & Pre-trained models & $M_{\Ff}$ & Pre-trained models & $M_{\Ff}$ & Ensemble & Pre-trained models & $M_{\Ff}$ & Ensemble  \\
        \midrule
        Fuse 2 RESNET & RESNET34, RESNET18 & RESNE34 & 21M, 11M & 21M & 32M & 1.78, 1.08 & 1.78 & 2.72 \\
        Fuse 4 RESNET & (RESNET34, RESNET18) x2 & RESNE34 & (21M, 11M) x2 & 21M & 64M & (1.78, 1.08) x2 & 1.78 & 5.30 \\
        Fuse 6 RESNET & (RESNET34, RESNET18) x3 & RESNE34 & (21M, 11M) x3 & 21M & 96M & (1.78, 1.08) x3 & 1.78 & 7.88 \\
        \bottomrule
    \end{tabular}
    }
    \label{table:inference_time}
\end{table*}

\subsection{Computational efficiency of model fusion}
\label{subsec:computational_efficiency}
In this section, we demonstrate that CLAFusion enhances the computational efficiency substantially over ensemble learning and knowledge distillation in all experiments. 

\vspace{0.5em}\noindent
Table~\ref{table:inference_time} illustrates the model size and inference time on the CIFAR10 test dataset of all models in Section~\ref{subsec:eff_init}. As discussed in Section~\ref{sec:introduction}, ensemble learning is computationally expensive because it stores and runs all pre-trained models during the inference stage. As the number of pre-trained models increases, memory consumption and inference time of ensemble learning grow up dramatically. In particular, the number of parameters multiplies by the same factor as the number of pre-trained models, increasing from 32M to 96M. Though rising at a lower speed, the inference time still increases by almost $190\%$, from 2.72 to 7.88. On the other hands, the fused model produced by CLAFusion remains compact no matter how many pre-trained models are fused.

\begin{table*}[t!]
    \centering
    \caption{Training time for finetuning and knowledge distillation in the teacher-student setting. The training time is the time taken in seconds to run one epoch of training on the whole train dataset.}
    \scalebox{0.65}{
    \begin{tabular}{cccccccccccc}
        \toprule
        Experiment & \multicolumn{3}{c}{Architecture} & \multicolumn{3}{c}{\# Params} & \multicolumn{3}{c}{Finetuning} &\multicolumn{2}{c}{Distillation} \\
        \cmidrule(lr){2-4} \cmidrule(lr){5-7} \cmidrule(lr){8-10} \cmidrule{11-12}
        & $M_A$ & $M_B$ & $M_{\Ff}$ & $M_A$ & $M_B$ & $M_{\Ff}$ & $M_A$ & $M_B$ & $M_{\Ff}$ & $M_B$ & $M_{\Ff}$  \\
        \midrule
        Teacher-student VGG & VGG13 doub & VGG11 half & VGG13 stud & 33M & 3M & 3M & 11.14 & 7.40 & 7.48 & 12.22 & 13.86 \\
        Teacher-student RESNET & RESNET34 & RESNET18 half & RESNET34 half & 21M & 3M & 5M & 31.73 & 7.50 & 12.05 & 36.39 & 41.36 \\
        \bottomrule
    \end{tabular}
    }
    \label{table:training_time}
    \vskip 0.1in
\end{table*}

\vspace{0.5em}\noindent
Table~\ref{table:training_time} compares the training time between finetuning and knowledge distillation in the teacher-student setting. It is undoubtedly that knowledge distillation runs much more slowly than finetuning, causing a rise of at least $65\%$ if using the same VGG initialization. In terms of RESNET architecture, the speedup is even more significant, with increases of $385\%$ and $243\%$ for $M_B$ and $M_{\Ff}$, respectively. The convincing reason is that in knowledge distillation the input data is fed in the forward direction through both teacher and student models while finetuning only runs forward inference for the student model whose size is much smaller than that of the teacher model.

\bibliography{arXiv_clafusion}
\bibliographystyle{abbrv}

\end{document}